\pdfoutput=1

\documentclass[11pt]{article}

\usepackage[]{emnlp2021}

\usepackage{times}
\usepackage{latexsym}
\usepackage{graphicx}
\usepackage{caption}
\usepackage{subcaption}
\usepackage{adjustbox}
\usepackage{booktabs}
\usepackage{comment}
\usepackage{siunitx}
\usepackage{multirow}
\usepackage[toc,page]{appendix}

\usepackage[T1]{fontenc}

\usepackage[utf8]{inputenc}

\usepackage{microtype}

\title{The Low-Resource Double Bind: An Empirical Study of Pruning for Low-Resource Machine Translation}

\author{Orevaoghene Ahia \\
  Masakhane NLP  \\
  \texttt{oreva.ahia@gmail.com} \\\And
  Julia Kreutzer \\
  Google Research  \\
  Masakhane NLP \\
  \texttt{jkreutzer@google.com} \\
  \\\And
  Sara Hooker \\
  Google Research, Brain \\
  \texttt{shooker@google.com} \\
  }

\begin{document}
\maketitle
\begin{abstract}
A ``bigger is better'' explosion in the number of parameters in deep neural networks has made it increasingly challenging to make state-of-the-art networks accessible in compute-restricted environments. Compression techniques have taken on renewed importance as a way to bridge the gap. However, evaluation of the trade-offs incurred by popular compression techniques has been centered on high-resource datasets. In this work, we instead consider the impact of compression in a data-limited regime. We introduce the term \emph{low-resource double bind} to refer to the co-occurrence of data limitations and compute resource constraints. This is a common setting for NLP for low-resource languages, yet the trade-offs in performance are poorly studied.\\
Our work offers surprising insights into the relationship between capacity and generalization in data-limited regimes for the task of machine translation. Our experiments on magnitude pruning for translations from English into Yoruba, Hausa, Igbo and German show that in low-resource regimes, sparsity preserves performance on frequent sentences but has a disparate impact on infrequent ones. However, it improves robustness to out-of-distribution shifts, especially for datasets that are very distinct from the training distribution. Our findings suggest that sparsity can play a beneficial role at curbing memorization of low frequency attributes, and therefore offers a promising solution to the low-resource double bind.
\end{abstract}

\section{Introduction}
Over the years, the size of language models have grown exponentially~\citep{2018Amodei,2020arXiv200705558T,bender_gebru_2021}. Additional parameters have improved quality on a variety of downstream NLP tasks, but drive up the cost of training~\citep{2014Horowitz,strubelL2019energy,patterson2021carbon} and increase the latency and memory footprint at inference time~\citep{warden2019tinyml, Samala_2018}.

Extending state-of-the-art language models to low-resource languages requires addressing what we term the \textit{low-resource double bind}. Low-resourcedness goes beyond mere data availability and reflects systemic issues in society~\cite{afocus, nekoto-etal-2020-participatory}. 
Classifications of languages with respect to ``resourcedness'' have focused on the relative availability of data ~\citep{zoph-etal-2016-transfer,joshi-etal-2020-state}, and the concentration of NLP researchers from these regions or the over-fitting of model design around a small set of high resource languages~\citep{cieri-etal-2016-selection,nekoto-etal-2020-participatory}. 

\begin{figure}
    \centering
    \includegraphics[width=0.8\columnwidth]{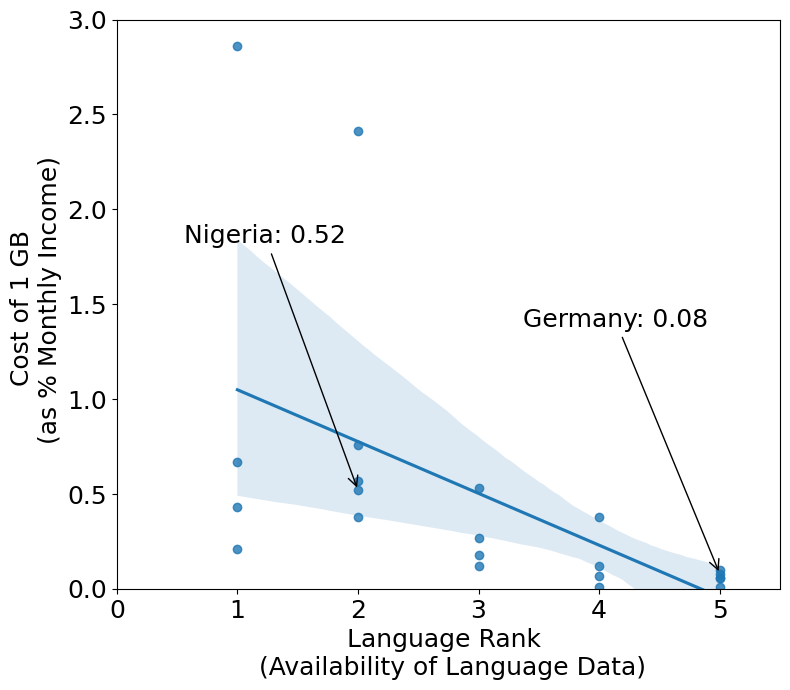}
    \caption{Cost of mobile data by country per language rank according to the taxonomy by~\citet{joshi-etal-2020-state}.}
    \label{fig:double_resource}
\end{figure}

Less well documented and explored is the over-indexing of low-resource languages in ecosystems which simultaneously present severe constraints of computational resource. In Fig.~\ref{fig:double_resource} we plot 22 languages grouped by the availability of labelled and unlabelled data as proposed by~\citet{joshi-etal-2020-state} against the cost of 1 GB of data as a percentage of monthly income. Each language is mapped to the country with the most speakers. The cost of data is a valuable proxy for the cost of access to technology in an ecosystem~\citep{oughton2021}. Here, this visibly co-occurs with the limitations in available data for different languages.

In computationally constrained environments, access to machine learning technology depends upon optimizing jointly for both model performance and compactness. Pruning and quantization are widely applied techniques for compressing deep neural networks prior to deployment, as compressed models require less memory, energy consumption and have lower inference latency \citep{2017Andre,8364435,sun_computation_sparse}. To-date, evaluating the merits and trade-offs incurred by compression have overwhelmingly centered on settings where the data is relatively abundant~\citep{2019arXiv190209574G,Li2020LearningLT,Hou2020DynaBERTDB,Chen2021EarlyBERTEB,Bai2020BinaryBERTPT,tessera2021gradients}.

In this work, we \emph{instead} ask how these design choices trade-off with performance in data-limited regimes typical of low resource languages. We conduct large scale experiments on Neural Machine Translation (NMT) models trained to translate between English and three low resource African languages (Yoruba, Igbo and Hausa) and one high resourced language (German). We compare performance across models independently trained to very different levels of sparsity --- ranging from 50 \% to 98 \% --- and evaluate performance on the original distribution, in addition to establishing sensitivity to distribution shift across multiple corpora.

Recent work restricted to the computer vision domain has found that sparse models with comparable top-line performance metrics diverge considerably in behavior on the long-tail of the distribution and are sensitive to distribution shifts ~\citep{hooker2020compressed,liebenwein}. Here, we rigorously characterize the impact of sparsity on learned decision boundaries in NMT. In addition to held-out set BLEU, we measure sub-group performance on sentences grouped by prototypicality and study generalization properties over test corpora with different out-of-vocabulary ratios. We also evaluate whether humans prefer translations from sparse or dense models.

Our contributions can be enumerated as follows:
\begin{enumerate}
\itemsep0em
     \item We introduce the term \textit{low-resource double-bind} and develop an extensive experimental framework to understand the impact of compression in a data-limited regime across 4 languages and 5 different data sets. 
    \item We find that models are \emph{tolerant of high levels of sparsity} while retaining BLEU performance and also human-judged translation quality. This holds until extremely high levels of sparsity (95\%--99\% of all weights removed) where a severe decline in BLEU is notable. 
    \item There is a more pronounced degradation when evaluation includes less frequent input patterns. On closer investigation, we find that \emph{sparsity disproportionately degrades performance on the long-tail of the data distribution}.
    \item Curbing memorization of the long-tail can provide unexpected benefits. In a data-limited regime, we find that \emph{sparsity benefits generalization to out-of-distribution corpora}.
\end{enumerate}

\paragraph{Implications of Our Work} Understanding the impact of compression on low-resource languages is key to making technology accessible and inclusive. Our work suggests that compression in these settings alters generalization in ways that can be beneficial and go beyond merely fulfilling deployment constraints. A challenge in low-resource NLP is that the existing publicly available corpora often come from very specific domains, such as missionary websites or translations of religious texts. These sources do not adequately reflect the reality of the potential applications of NLP technologies, and are rarely sufficient for deployment~\citep{tigrinya,tico,congolese}. Thus, a task of great interest is establishing what model design choices can lead to generalization properties that extend beyond the immediate task at hand. Our work suggests that sparsity can play an important role in aiding generalization by curbing the memorization of rare long-tail instances. 

\section{Methodology}

Addressing the low-resource double bind requires a careful setup of experiments to reflect the realities of low-resource translation. In particular, we want to control the effects of (1) network sparsity, (2) training data size, (3) target language, and (4) domain shifts. 

In this work we focus on pruning, a widely favored compression technique due to remarkably high levels of compression that can be achieved while retaining top-line performance \citep{tgale_shooker_2019}. Pruning typically involves three separate stages: 1) training a dense model, 2) progressively removing a subset of weights estimated to be unimportant, and 3) continuing to train the smaller sparse network for a certain number of steps to recoup performance ~\citep{248452,blalock2020state}.
Pruning is the subject of considerable research and numerous techniques have been proposed, which differ in how weights are identified for removal and the schedule for introducing sparsity/allowing recovery ~\citep{Cun90optimalbrain,1993optimalbrain,Strom97sparseconnection,2017l0_reg,2016abigail,evci2019rigging, 2017Narang}.  
The development of specialized software kernels has enabled the acceleration of sparse networks on traditional hardware  \citep{gale2020sparse,elsen2019fast,sparse_tensor_core} with new generations of hardware directly facilitating sparse training~\citep{sparse_tensor_core}. 

State of art pruning techniques can achieve a far higher level of compression and performance than simply using a smaller dense network \citep{to-prune-or-not, rethinking_model_size_li}. In our setting, a 90\% sparse base transformer greatly outperforms a tiny dense one across all the languages despite having a fraction of the parameters (14M vs 4.6M) (Appendix Table~\ref{tab:parameter_count}).

\begin{figure*}[ht!]
	\centering
	\vskip 0.15in
\begin{small}
\begin{sc}
\begin{subfigure}{0.28\linewidth}
		\centering
    	\includegraphics[width=0.99\linewidth]{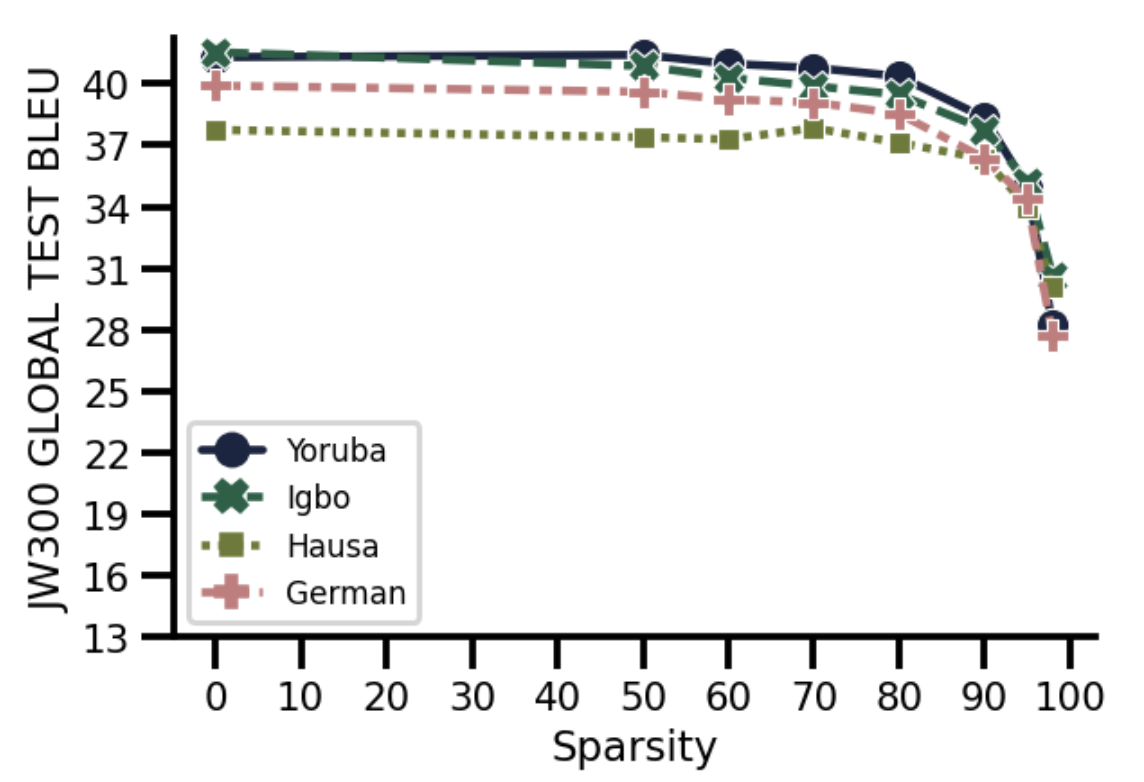}
    		\caption{Global Test \texttt{Full}}\label{subfig:global_test_full}
	\end{subfigure}	
	\hspace{1em}%
\begin{subfigure}{0.28\linewidth}
	\centering	\includegraphics[width=0.99\linewidth]{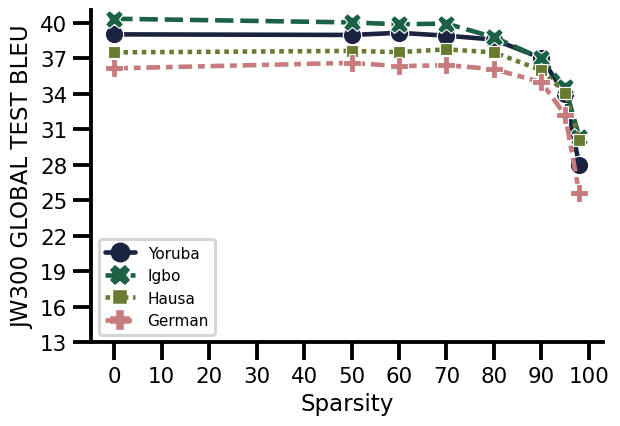}
		\caption{Global Test \texttt{Limited}}\label{subfig:global_test_limited}
\end{subfigure}
\hspace{1em}%
	\begin{subfigure}{0.28\linewidth}
	\centering
	\includegraphics[width=0.99\linewidth]{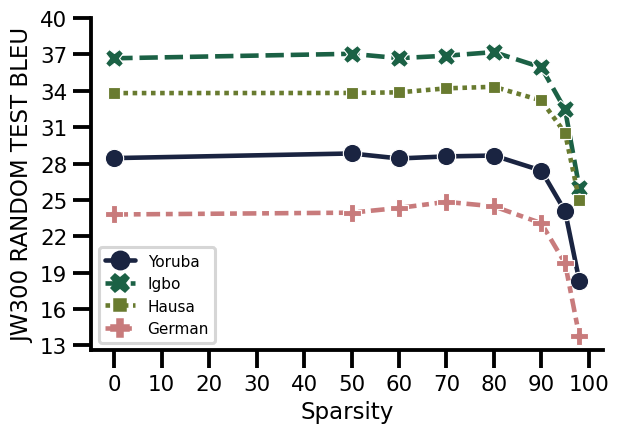}
		\caption{Random Test \texttt{Limited}}
\end{subfigure}
 	 \\
 \begin{subfigure}{0.28\linewidth}
		\centering	\includegraphics[width=0.99\linewidth]{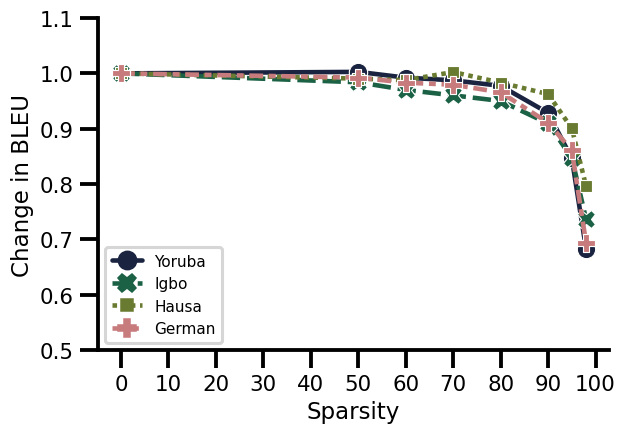}
\caption{Global Test \texttt{Full}}
\end{subfigure}
	\hspace{1em}%
   \begin{subfigure}{0.28\linewidth}
		\centering	\includegraphics[width=0.99\linewidth]{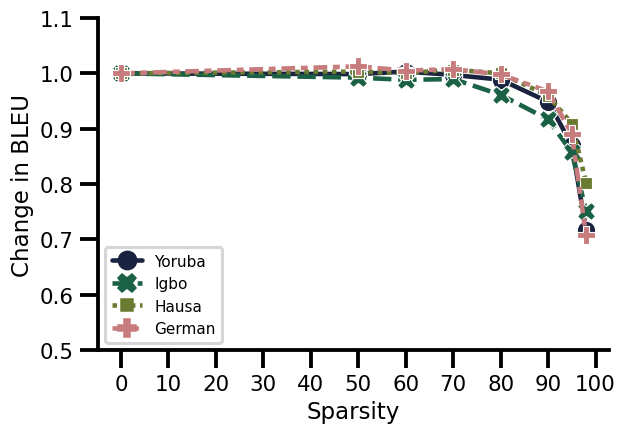}
    		\caption{Global Test \texttt{Limited}}
	\end{subfigure}
	\hspace{1em}%
	\begin{subfigure}{0.28\linewidth}
		\centering	\includegraphics[width=0.99\linewidth]{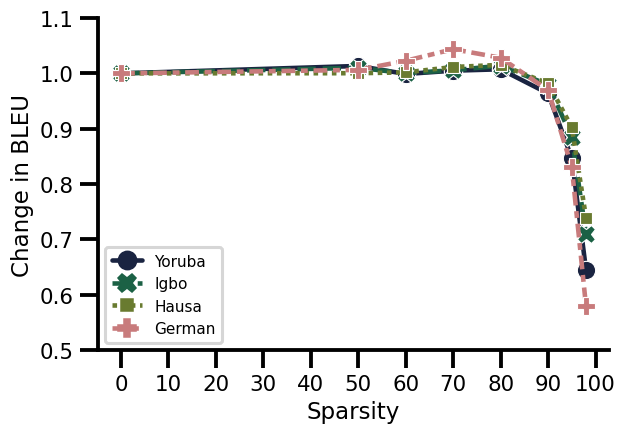}
    		\caption{Random Test \texttt{Limited}}\label{fig:randomtest}
	\end{subfigure}
	\end{sc}
	\end{small}
	\caption{
	   Impact of pruning on BLEU performance across languages, sparsity levels and training data regimes. We evaluate test set performance on both a \texttt{Global} test set designed around common phrases to allow comparability between data corpus, and a \texttt{Random} test set with sentences sampled at random. \textbf{Top row:} Absolute change to BLEU and by test set and sample size \textbf{Bottom row:} 
	Change in BLEU relative to the dense (0\% sparse) model.}
	\label{fig:degradation}
\end{figure*}

\subsection{Magnitude Pruning}\label{magnitude pruning technique}

We use magnitude pruning ~\citep{to-prune-or-not} to introduce sparsity across all experiment variants. It consistently achieves comparable or better results than competing state of art approaches on large scale benchmarks of computer vision and language models ~\citep{2019arXiv190209574G} and is widely used in practice due to the ease of implementation. Magnitude pruning estimates weight importance as the absolute weight value, and removes the weights with lowest magnitude according to a pre-specified schedule which determines the interval of training steps and frequency between begin and end step across which sparsity is introduced. 

Magnitude pruning allows for the pre-specification of desired sparsity such that we can train models from random initialization to precise levels of end sparsity. We carry out extensive experiments and train networks independently for each language to end sparsity of 0--98 where 98\% designates a network with 98\% of the weights removed by the end of training. 0\% is a dense network (no weights removed).

\subsection{Languages}
We validate the effectiveness of magnitude-based pruning method in NMT models trained to translate from English into German (\texttt{de}), Yoruba (\texttt{yo)}, Igbo (\texttt{ig}) and Hausa (\texttt{ha}). While German as a high-resource language serves as a point of comparison to previous works, Yoruba, Igbo and Hausa represent three of the highest-resource African languages with (near-)sufficient resources for reliable MT experimentation, i.e. multiple publicly-available parallel corpora. \citet{joshi-etal-2020-state} classify Yoruba and Hausa as ``Hopeful'' in terms of available NLP resources and research, whereas Igbo is slightly lower-resourced and classified as ``Scraping-by''. All constitute important test beds for developing technologies that improve treatment of low-resource technologies, since they each have more than 50 million native speakers. Yoruba and Igbo belong to the Niger-Congo language family and use diacritics that pose challenges for text-based NLP~\citep{orifeDiacritics,OkwuGbe}. Hausa is a Chadic language which is part of the Afroasiatic phylum. It features complex pluralization and agglutination.

\begin{table}
\centering
\resizebox{1.0\columnwidth}{!}{
\begin{tabular}{l| c |c c c c c c}
\toprule
& Training & \multicolumn{5}{c}{Distribution Shift Test} \\
 & \textbf{JW300} & \textbf{Gnome} & \textbf{Ubuntu} & \textbf{Flores} &\textbf{ParaCrawl} & \textbf{Tanzil} & \textbf{Tatoeba} \\  
\midrule 
\texttt{de} &   1.9M   &   5,963    &  11,161  & 1012  &   2,000  & 2,000 & 10,145 \\
\texttt{yo}  &   414.0k     &   1,467    &  120   & 1012    &   -     & -  & -  \\
\texttt{ig}    &   414.9k     &   3,173    &   608  & 1012    &   2,000  &  -  & -  \\
\texttt{ha}   &   211.9k     &   998     &   219    & 1012  &   2,000   & 2,000  & - \\ 
\bottomrule
\end{tabular}%
}
\caption{Number of sentences in each parallel corpora we evaluate. For ParaCrawl and Tanzil, we sample $2000$ sentences from the full dataset.}
\label{tab:data}
\end{table}

\subsection{Training and Test Data}
\paragraph{JW300} Training data for all languages is obtained from the JW300 parallel corpus ~\citep{agic-vulic-2019-jw300}, since it is the largest source of data that covers all languages we evaluate. It comprises more than 300 languages of which 101 are African, and is collected from religious magazines by Jehovah’s Witnesses (JW) published on \url{jw.org}.

\paragraph{Pre-processing} Parallel sentences are tokenized and encoded using BPE \citep{sennrich-etal-2016-neural}, resulting in a shared vocabulary of \SI{4096} tokens. Sentences are batched together with a maximum sequence length of \SI{64}{}. For each training batch, the approximate number of source/target tokens is \SI{2048}{}. We compute detokenized and case-sensitive BLEU using a helper script in tensor2tensor \citep{tensor2tensor} equivalent to SacreBLEU \citep{post-2018-call}.  

\paragraph{Full vs limited data regime} For our experiments, we train on these datasets in two settings: First, with all data available for each of the languages, sizes listed in Table~\ref{tab:data}. In this setting, the dataset sizes range from 212k for Hausa to 1.9M for German.
Our second setting holds constant the amount of data available by sampling a uniform number of sentence pairs for each language. We randomly sample 200k sentences from the train set of each language, limited by the smallest corpus Hausa which consists of approximately 210k sentences. We refer to these settings in experiment discussion as \texttt{Full} and \texttt{Limited}. 

\paragraph{Validation \& testing} 
The need for multiple test sets to capture performance on a variety of downstream conditions has already been recommended by recent work~\citep{sogaard-etal-2021-need,lazaridou2021pitfalls}. 
The JW300 test sets were constructed and released by ~\citet{nekoto-etal-2020-participatory} to contain the most frequent sentences in the JW300 corpus across African languages and were filtered from the training corpus. This construction ensures that test sets across languages contain similar content, which leads to increased comparability. However, this cross-lingual selection may introduces a bias towards frequent sentences, and under-represents language-specific outliers. 

Only measuring performance on frequent sentences across languages may be a particular concern in evaluating the impact of sparse models, as prior work has shown that the introduction of sparsity disproportionately impacts the long-tail of the data distribution ~\citep{hooker2020compressed,hooker2020characterising}. 
To capture possible disparate impact on the long-tail, we also sample at random from the remainder of the data to craft a secondary test set (as has been done for validation). In the results section, we refer to the ~\citet{nekoto-etal-2020-participatory} test data as the \texttt{Global} test set and random sample as the \texttt{Random} test set. A comparison of differences in performance between \texttt{Global} and \texttt{Random} test sets provides insights into how sparsity impacts generalization performance on text which is common relative to a more typical Zipf distribution with long-tail features~\citep{zipf1999psycho}.

\subsection{Sensitivity to Distribution Shift}

We select corpora which differ from the training distribution in both domain (ranging from everyday sentences to technical documentation), sentence length and OOV rate (ranging from 2.68\% to 20.42\%). Given these large deviations in statistics from the JW300 training corpus, our expectation is \emph{not} that the model preserves performance but rather to understand the sensitivity of sparse models to distribution shift \textit{relative} to dense models.

\begin{figure}
    \centering
    \includegraphics[width=\columnwidth]{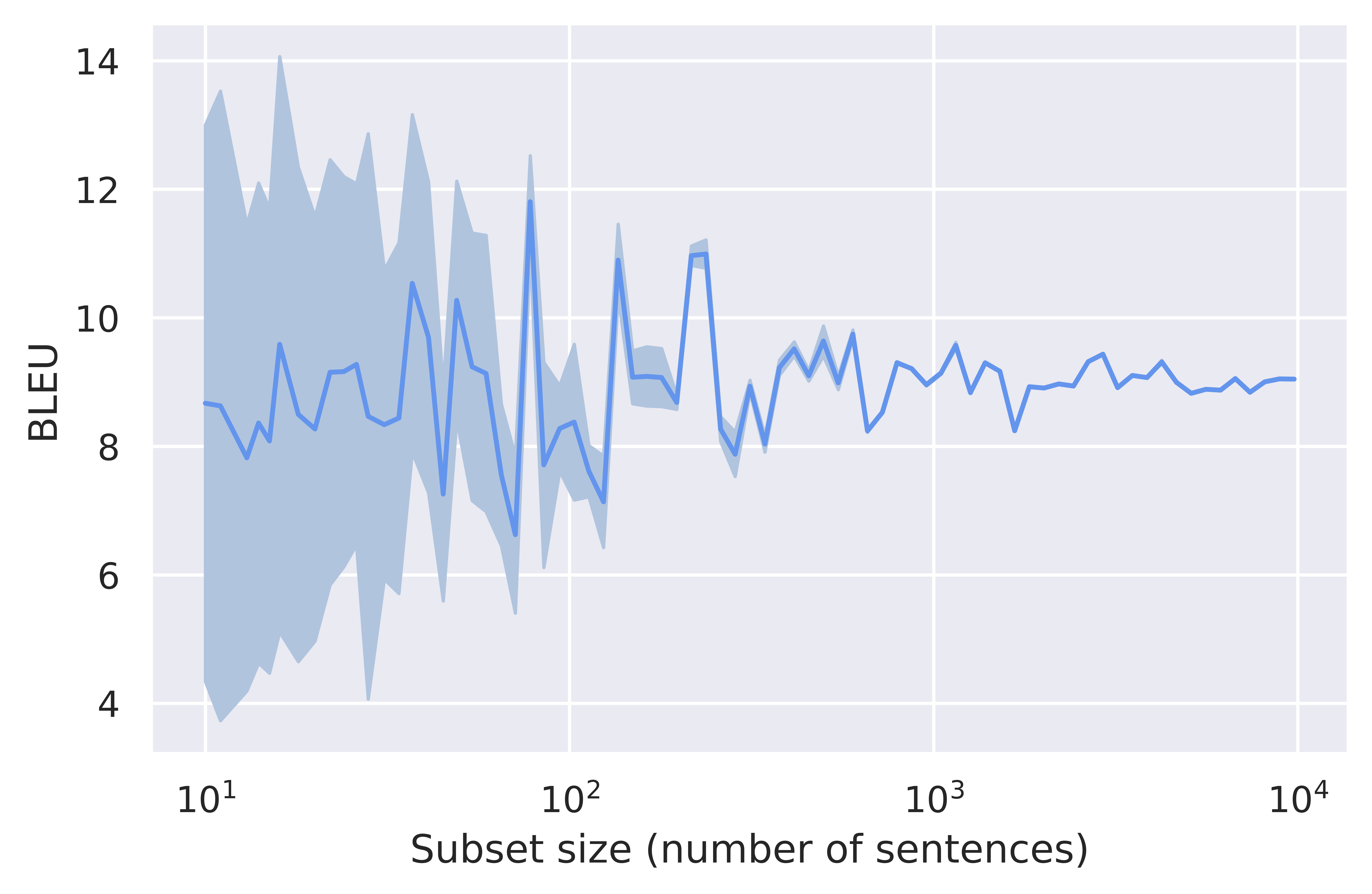}
    \caption{Mean BLEU scores (shaded: $\pm$ standard variation) for the dense en-de models on subsets of the Tatoeba data.}
    \label{fig:subset}
\end{figure}

Our selection of corpora is also guided by the size of public data sets that cover Yoruba, Hausa, Igbo and German.  When the test set is small, reliability in BLEU scores between models and inferred conclusions may be compromised~\citep{card-etal-2020-little}. To estimate the impact that limitation in test size can have on our results, we simulate the variability of BLEU under different amounts of test data in Figure~\ref{fig:subset}. As can be seen, a sample size of at least $100$ reasonably reduces variance in BLEU, so we only investigate out-of-distribution sensitivity with datasets of at least that size.

The domains of the datasets can be  characterized below. Statistics for corpus sizes are given in Table~\ref{tab:data}, and out-of-vocabulary rates (OOV) and average source lengths in Table~\ref{tab:oov}.

\subsection{Datasets evaluated}

The domains of each dataset considered are characterized below. Additionally, we include statistics for 1) corpus size in Table \ref{tab:oov}, and 2) out-of-vocabulary rates (OOV) and average source lengths in Table~\ref{tab:data}:
\begin{itemize}
\item \textbf{Gnome} is a dataset in the technical domain that contains
187 languages pairs derived from the translation of GNOME documentation.\footnote{\url{https://www.gnome.org/}} The size of test sets for this corpus ranges between 998 (Hausa) and 5,963 (German). ``Sentences'' are often entities or phrases, with an average length of only 6-9 tokens.
\item \textbf{Ubuntu} is a dataset in the technical domain. It consists of 42 languages pairs generated from translating localization files of the Ubuntu OS.\footnote{\url{https://ubuntu.com/}}  The size of test sets for this corpus ranges between 120 (Yoruba) and 11,161 (German), and it shows similar length statistics to GNOME.
\item \textbf{Tanzil} is a religious dataset with 42 language pairs. It is a collection of Quran translations compiled by the Tanzil project.\footnote{\url{https://tanzil.net/}} We sample 2000 sentences for both German and Hausa, which have an average length of 23 tokens, being slightly longer than the average JW300 training sentence.
\item \textbf{ParaCrawl} is a dataset obtained from mining the web for parallel sentences~\citep{banon-etal-2020-paracrawl}. v8 covers 41 mostly European languages, but a pre-release of Igbo and Hausa allowed evaluation here. \footnote{\url{https://bit.ly/3f7WfVI}} The crawled websites for Hausa and Igbo are largely religious but some also publish news. We sample 2000 sentences for Hausa, Igbo and German with an average length of 22 tokens.
 \item \textbf{Tatoeba} is a crowdsourced dataset of short sentences concerning every day life translated by users of~\url{https://tatoeba.org/}. We only report Tatoeba results for German as this is the only corpus with more than 100 sentences. Tatoeba sentences have similar length to Gnome and Ubuntu, but are full sentences.
\item \textbf{Flores} is a multi-domain dataset containing professional translations of sentences extracted from English Wikipedia in 101 languages~\citep{flores}. The size of test sets released for this corpus is 1012 sentences across all languages with similar length to Tanzil and Paracrawl. 
\end{itemize}

Our choice of datasets is guided by a desire to capture datasets with different degrees of difference from the original training corpus. JW300 is a religious dataset, so one could expect more overlap with both \textbf{ParaCrawl} and \textbf{Tanzil} and far less with \textbf{Ubuntu} and \textbf{Gnome} which are both technical writing to document the use of a technology. We include \textbf{Flores} which covers a variety of different domains and finally \textbf{Tatoeba} for completeness, as a more general dataset consisting of everyday sentences.

\begin{table*}[]
    \centering
    \resizebox{\textwidth}{!}{

    \begin{tabular}{l|ccc|ccc|ccc}
    \toprule
  &  \multicolumn{3}{c|}{\textbf{Training}} & \multicolumn{3}{c|}{\textbf{\texttt{Global} test}} & \multicolumn{3}{c}{\textbf{\texttt{Random} test}}\\
         & Avg Len & Dense  & 90\% Sparse  & Avg Len & Dense & 90\% Sparse  & Avg Len & Dense & 90\% Sparse \\
    \midrule
        Low & 33.01 & 62.04  &  24.93 & 23.58 & 31.41 & 29.45 & 32.73 & 22.35 & 23.08\\
        Mid & 18.26 & 80.09  & 26.82 & 14.03 & 35.37 & 34.68 & 17.53 & 23.99 & 23.58\\
        High & 8.99 & 78.91 & 28.58 & 9.86 & 48.56 & 48.05 & 9.09 &25.47 & 24.86\\
    \bottomrule
    \end{tabular} %
    }
    \caption{BLEU for different sets split according to sentence typicality for German, which is defined as average token log-frequencies in the training corpus ($F_S$ in ~\citep{raunak-etal-2020-long}).}
    \label{tab:typicality}
\end{table*}

\begin{table*}[]
    \centering
    \resizebox{\textwidth}{!}{

    \begin{tabular}{l|cc|cc|cc|cc|cc|cc|cc|cc}
    \toprule
       &  \multicolumn{2}{c|}{\textbf{JW300 \texttt{Global}}} & \multicolumn{2}{c|}{\textbf{JW300 \texttt{Random}}} &  \multicolumn{2}{c|}{\textbf{Tanzil}} & \multicolumn{2}{c|}{\textbf{Tatoeba}} & \multicolumn{2}{c|}{\textbf{ParaCrawl}} & \multicolumn{2}{c|}{\textbf{Gnome}} & \multicolumn{2}{c|}{\textbf{Ubuntu}} & \multicolumn{2}{c}{\textbf{Flores}}\\
        & OOV & Len & OOV & Len & OOV & Len & OOV & Len & OOV & Len & OOV & Len & OOV & Len & OOV & Len\\
                \midrule

         \texttt{de} & 0.25 & 15.81 & 0.66 & 19.76 &  2.68 & 22.48 & 4.89 & 8.78 & 9.86 & 20.09 & 12.37 & 8.09 &16.64 & 5.56 & \multirow{4}{*}{15.53} & \multirow{4}{*}{21.64}\\
         \texttt{ha} & 0.26 & 16.28 & 0.37 & 18.72 &  4.95 & 22.86 & 7.00 & 7.76 & 3.39 & 24.67 & 15.22 & 9.81 & 20.42 & 7.83 \\
         \texttt{ig} & 0.30 & 15.98 & 0.50 & 18.58 & - & - & 12.10 & 6.89  & 6.95 & 20.91 & 14.19 & 6.99 & 13.99 & 6.81\\
         \texttt{yo} & 0.24 & 15.98 & 0.56 & 18.77 & - & - & 9.05 & 5.69 & - &- & 16.66 & 6.36 & 13.55 & 6.46\\
    \bottomrule
    \end{tabular}%
    }
    \caption{Out-of-vocabulary rates (OOV, \%) and average source lengths (Len) for different test set sources.} 
    \label{tab:oov}
    	\vspace{-0.2cm}
\end{table*}

\subsection{Architecture and Training}\label{sparse_transformer}

We train transformer models~\citep{vaswani2017attention} for each NMT task with a modified version of the tensor2tensor library~\citep{tensor2tensor} from~\citep{tgale_shooker_2019}. The transformer base model consists of 60M parameters, with $31\%$ of the parameters in the attention layers, and $41\%$ in the position wise feed-forward layers. Training hyperparameters are detailed in Appendix Section~\ref{sec:hyper-training}. We release our code here \url{https://github.com/orevaahia/mc4lrnmt}.

Throughout training we introduce sparsity of levels percentages [0, 50, 60, 70, 80, 90, 95, 98] using magnitude pruning~\citep{to-prune-or-not}. All fully-connected layers and embeddings making up 99.87\% of all of the parameters in the model are considered for sparsification.
The tuning of pruning hyper-parameter is described in Appendix Section~\ref{sec:hyper}.

\subsection{Human Evaluation: Dense vs Sparse}\label{sec:human}
We complement automatic BLEU evaluation with a human evaluation study to compare the translation quality of dense and sparse models. We elicit absolute ratings on a 6-point scale for 500 pairs of differing translations of the JW300 \texttt{Global} and \texttt{Random} test set on a crowd-sourcing platform.

\begin{figure*}[th!]
	\centering
	\vskip 0.15in
\begin{small}
\begin{sc}
	\begin{subfigure}{0.23\linewidth}
		\centering
		
    	\includegraphics[width=0.99\linewidth]{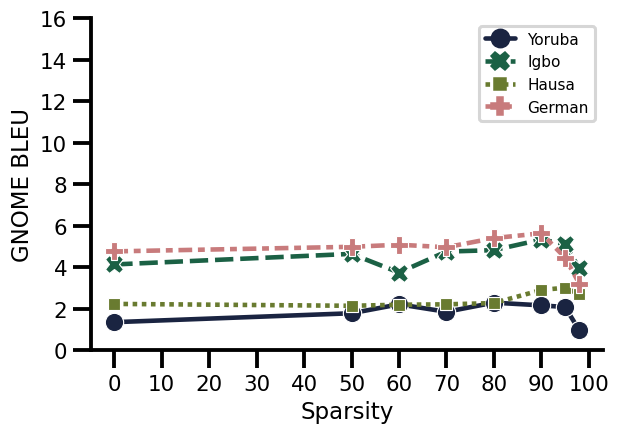}
    		\caption{Gnome} 
	\end{subfigure}
	\begin{subfigure}{0.23\linewidth}
		\centering
		
    	\includegraphics[width=0.99\linewidth]{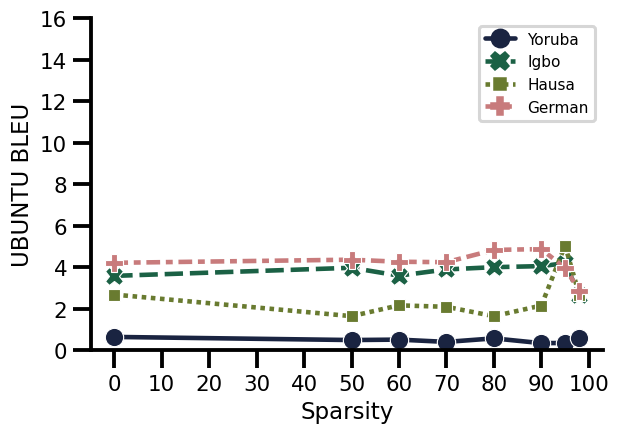}
    		\caption{Ubuntu}
	\end{subfigure}	
		\begin{subfigure}{0.23\linewidth}
		\centering
    	\includegraphics[width=0.99\linewidth]{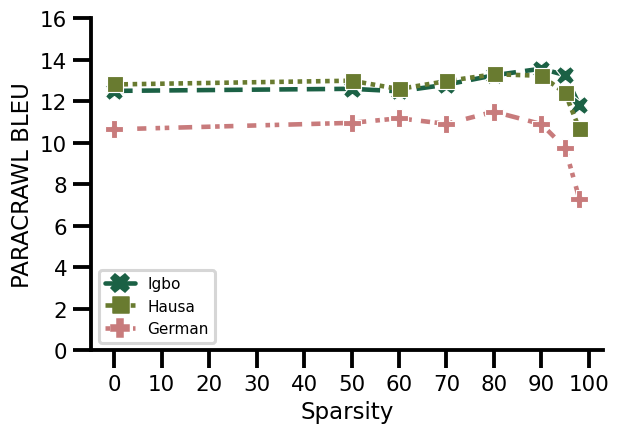}
    		\caption{ParaCrawl}
	\end{subfigure} 
	\begin{subfigure}{0.23\linewidth}
		\centering
    	\includegraphics[width=0.99\linewidth]{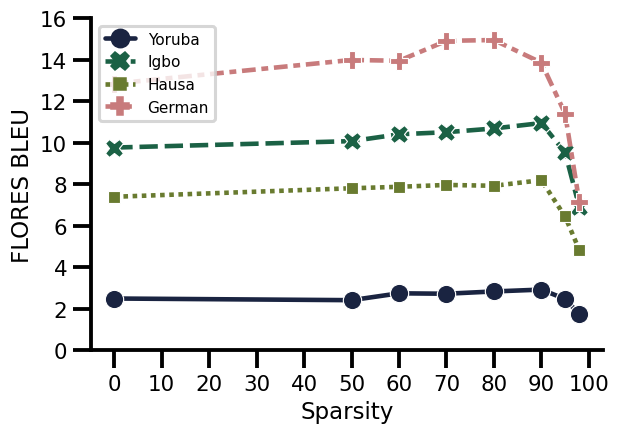}
    		\caption{Flores} 
	\end{subfigure} \\	
	\begin{subfigure}{0.30\linewidth}
		\centering
    	\includegraphics[width=0.99\linewidth]{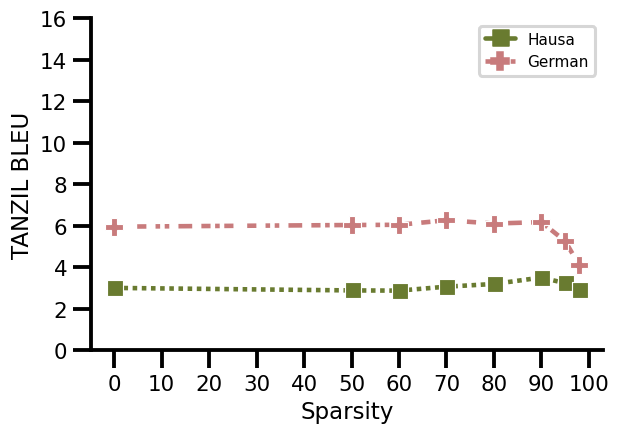}
    		\caption{Tanzil}
	\end{subfigure}	
		\begin{subfigure}{0.30\linewidth}
		\centering
        \includegraphics[width=0.99\linewidth]{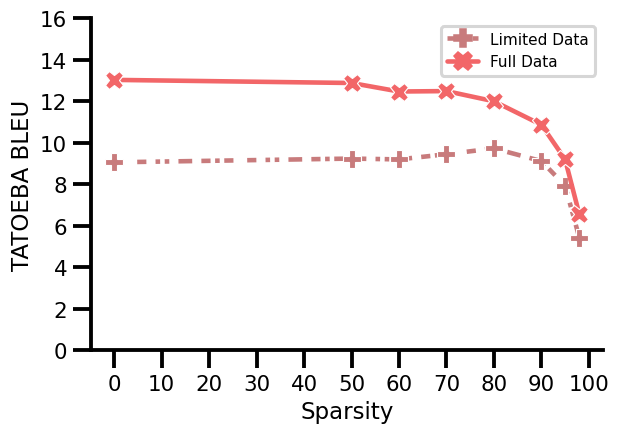}
    		\caption{Tatoeba: \texttt{Limited} vs \texttt{Full}}
    		\label{fig:de_full_sample}
    		\end{subfigure}	
    			\begin{subfigure}{0.30\linewidth}
		\centering	\includegraphics[width=0.99\linewidth]{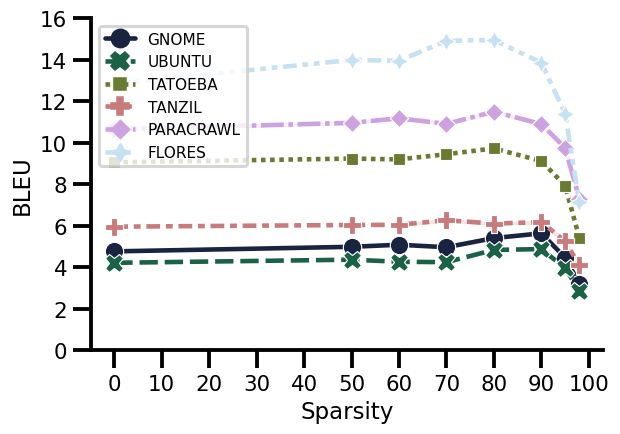}
    		\caption{All (German)}
	\end{subfigure}
	\end{sc}
	\end{small}
	\caption{
       Robustness to distribution shift at different levels of sparsity for models trained in a data-limited regime (\texttt{Limited}). For Tatoeba with only German we compare performance for a model trained on \texttt{Full} is added for comparison.
	}
	\label{fig:sensitivity}
	\vspace{-0.5cm}
\end{figure*}

\section{Results}

\paragraph{Sparsity BLEU trade-off} In Figure~\ref{fig:degradation}, we can see that models are tolerant of moderate to high levels of sparsity (50\% - 80\%) while retaining BLEU performance relative to dense. Between 50\% and 80\% sparsity, any degradation is minimal as sparse performance remains at 95\% or more of dense performance for all languages. Hausa even has a slight exception where pruning 70\% or 80\% of the model parameters performs on par with the baseline or even better. However, for both \texttt{Global} and \texttt{Random} test sets, there is a noticeably sharp degradation in BLEU when progressing to extremely high sparsity levels of 90\% and beyond. 

\paragraph{Long-tail test set} Translation quality on the \texttt{Global} and \texttt{Random} test sets differs considerably. We control for data size by comparing on the same \texttt{Limited} datasets. Languages perform within a narrow band of comparable BLEU for \texttt{Global}, with the degradation in BLEU at higher levels of sparsity occurring at a similar rate across languages. In contrast, absolute BLEU scores on \texttt{Random} are noticeably lower at both dense and all sparsity levels, coupled with a far wider spread of BLEU between languages. This suggests that \emph{a low data regime disproportionately impacts translation quality on the long-tail} and that the \texttt{Random} set is a more discriminative evaluation protocol. 
When we compare relative differences of sparse models to dense, we can see that relative to \texttt{Global}, there is sharper degradation in \texttt{Random} under high sparsity (90\%+). However, with mid-level sparsity, the quality of the dense model is maintained or even slightly outperformed (German) on all test sets.

\paragraph{Learning prototypical instances is less sensitive to data size} In Figure~\ref{fig:degradation}, it is noticeable that performance on the \texttt{Global} test set, does not vary noticeably between the \texttt{Limited} (\ref{subfig:global_test_limited}) and \texttt{Full} (\ref{subfig:global_test_full}) training setting. This is surprising given the large difference in training corpus size for many of the languages (for German 1.9 M in \texttt{Full} vs 200,000 in \texttt{Limited}).

Additionally, even when restricting attention to the \texttt{Full} training setting, the ranking of absolute BLEU scores on the \texttt{Global} test set does not appear to be sensitive to the size of the training corpus, as Igbo (414.9K) and Yoruba (414.0K) achieve nominally higher BLEU on \texttt{Global} than German (1.9M) despite having only a fifth of training data in the \texttt{Full} setup. This suggests that learning a representation for the most frequent patterns in the dataset does not require a substantial amount of data.

\paragraph{Data size for long-tail patterns} In contrast, learning a good representation for the long-tail appears to be far more sensitive to the size of the training corpus. In Figure~\ref{fig:de_full_sample}, we compare the OOD performance on Tatoeba of the \texttt{Full} vs \texttt{Limited} model trained on German. Here, we are evaluating performance on a dataset with a much higher OOV ratio of $2.68\%$. Here, on a dataset with more rare instances and distribution shift, the amount of training data makes a larger difference. Across all levels of sparsity the model trained on \texttt{Full} generalizes better than the \texttt{Limited} model.

\paragraph{Do humans notice the difference between dense and sparse models?}
Human annotators rate test translations of the dense model and 90\%-sparsity model as described in Section~\ref{sec:human}.
Table~\ref{tab:ratings} reports the average ratings (1-6 scale, the higher the better) for both types of models across languages for both in-domain test sets.
The ratings reveal that there is no clear preference of dense over sparse model outputs across languages and sets. For German the sparse model scores 0.1 lower on average for both test sets. For Igbo and Hausa sparse scores slightly higher on the \texttt{Global} set, but this gain is lost on \texttt{Random}. Hausa receives the nominally highest ratings on both test sets, but we note that 
raters might not be well calibrated across languages, and test sets are not completely identical. Nevertheless, the \texttt{Random} translations score consistently lower than the \texttt{Global} translations, indicating that the quality loss on long-tail examples was indeed noticeable.
All in all, \emph{the roughly 2-8\% drop in BLEU that occurred through pruning at 90\% is not negatively affecting human-judged translation quality in any of the studied languages}, which is a promising finding for the deployment of such models in practice. However, this evaluation is oblivious of effects like translation biases~\citep{edunov-etal-2020-evaluation} that could be caused by less memorization. 

\paragraph{How sensitive are sparse models to distribution shift?} \label{sec:distribution_shift_explanation}
Figure~\ref{fig:sensitivity} shows the absolute change in BLEU when evaluating the \texttt{Limited} models on the out-of-distribution datasets. 

We find that across dense and sparse models, degradation in performance is sensitive to OOV rates and difference in sentence lengths, with the most pronounced degradation on Tanzil (longer average sentence length), Ubuntu and Gnome (technical domains with far higher OOV rates of 12--20\%). The transfer to ParaCrawl was the most successful across languages. For Flores, we don't see a uniform performance across all languages. Our results show that the transfer to all languages but Yoruba is almost similar to that of Paracrawl. 

One trend that is consistent across languages and domains is an increase in quality around 70\%--95\% sparsity (even more visible when plotting relative change in Figure~\ref{fig:relative_degradation_distribution2} in the Appendix).
As a result, 90\% sparse models outperform their baselines across all languages under the \texttt{Limited} condition. This means that \emph{increased sparsity during training with limited data has a beneficial influence for out-of-distribution generalization}. With larger amounts of training data---in the case of German (\texttt{Entire}) a factor of 10--- however, this relative advantage is lost (Figure~\ref{fig:de_full_sample}). This finding is highly relevant for the reality of low-resource languages, where training data is limited and often available only for a narrow domain, so strong generalization is essential.

\paragraph{Does sparsity curb memorization?} The results on OOD generalization are surprising, as it suggests that in a low-data regime \textit{less capacity rather than more can aid generalization}. It is worth placing these results in the context of recent work in computer vision that has found that sparsity curbs memorization of the long-tail of the training distribution \citep{hooker2020compressed,hooker2020characterising}. Impeding memorization constrains model learning to the most general high level features, rather than atypical low-frequency features and noisy examples which typically require memorization and are less likely to generalize across other tasks \citep{brown2020memorization,NEURIPS2020_1e14bfe2}. In the setting of low-resource languages, the training corpus is often highly restricted to a narrow domain such as religion. Here, it may be more likely the long-tail will contain specialized artefacts and noise that do not generalize.

To explore this further, in Table~\ref{tab:typicality} we look at training and test performance grouped by sentence typicality for German (typicality measured as per \citep{raunak-etal-2020-long}). At training time, the dense model evidences clear overfitting and outperforms the sparse on low, mid and high typicality. The difference in relative performance on sentences of low typicality is striking (62.04 dense vs 24.93 sparse BLEU), confirming that capacity aids memorization of the long-tail. However, at test time the dense memorization of a specialized training set exhibits a negative knowledge transfer cost relative to sparse. On the \texttt{Random} test set, sparse in fact slightly outperforms relative to dense on low typicality. Both Table~\ref{tab:typicality} and the OOD result in Figure~\ref{fig:sensitivity} show that sparsity has an important role to play in data limited regimes at curbing memorization of rare artefacts that do not generalize.

\begin{table}[t]
    \centering
    \resizebox{0.9\columnwidth}{!}{
    \begin{tabular}{l|cc|cc}
    \toprule 
        & \multicolumn{2}{c|}{\textbf{\texttt{Global}}} & \multicolumn{2}{c}{\textbf{\texttt{Random}}} \\ 
       & Dense & 90\% Sparse & Dense & 90\% Sparse \\
        \midrule
        \texttt{de} & 4.17 & 4.06 & 3.80 & 3.66 \\
        \texttt{yo} & 3.66 & 3.66 & 3.51 & 3.51 \\ 
        \texttt{ig} & 3.87 & 3.96 & 3.81 & 3.85 \\ 
        \texttt{ha} & 4.77 & 4.85 & 4.53 & 4.53\\ 
    \bottomrule
    \end{tabular}%
    }
    \caption{Average human ratings of 500 test set translations comparing \textbf{dense} and \textbf{90\%-sparse} models (\texttt{Limited}) on \texttt{Global} and \texttt{Random} JW300 test sets.} 
    \label{tab:ratings}
    	\vspace{-0.2cm}
\end{table}

\section{Related Work}

\paragraph{Compression techniques for NMT} There has been recent works on compressing recurrent neural networks (RNN) and transformer networks for NMT ~\citep{2019arXiv190209574G,narang2017exploring, see-etal-2016-compression,zhang-etal-2017-towards,rethinking_model_size_li}. With the exception of a proof-of-concept experiment with RNN on a small-scale English-Vietnamese translation task~\citep{see-etal-2016-compression}, all of the works above focus on compressing models trained on large data sets, and exclude African Languages.

To the best of our knowledge, our work is the first to apply pruning methods to train transformer NMT models on low-resourced data, and on African languages with different syntactic and morphological features distinct from English. Moreover, all of the above works rely solely on automatic evaluation metrics, and do not qualify translation quality using human annotation or sensitivity to different distribution shifts.

\paragraph{Optimized training for low-resource NMT} 
\citet{sennrich-zhang-2019-revisiting} find that hyperparameter tuning is essential for NMT on low-resource data, such as the depth or regularization of the network. \citet{duh-etal-2020-benchmarking} highlight the importance of hyper-parameter tuning for NMT for Somali and Swahili. \citet{fadaee-etal-2017-data,sennrich-zhang-2019-revisiting} and \citet{xu-etal-2020-dynamic} explore tailored augmentation and curriculum learning strategies for data-limited regimes. \citet{sennrich-zhang-2019-revisiting} additionally assume limitations to compute at training time when modeling Somali and Gujarati. In contrast, in our work, we consider the impact of resource constraints present at inference time/deployment of a model.

\paragraph{Transformer size for low-resource NMT} More relevant to our work are works that have evaluated transformers of different sizes in the light of low-resource translation. \citet{Biljon2020OnOT} investigate the effect of transformer depth on low-resource translation for three South-African languages. \citet{murray-etal-2019-auto} study auto-size feed-forward and attention layers in transformers for low-resource translations of Hausa, Tigrinya, and Arabic, and find BLEU and efficiency improvements with smaller models. \citet{tapo-etal-2020-neural} succeed in training a smaller transformer model for Bambara with as few as 10k examples, but find only limited generalization under distribution shift ~\citep{Tapo2021}. 

In contrast to these works, we study generalization at different levels of sparsity. Pruning is a more precise experimental framework to understand the relationship between capacity and generalization because we can exactly vary the sparsity in a range between 0\% and 100\% controlling for the same architecture. Pruning also achieves far higher levels of compression in terms of the number of parameters relative to substitutes evaluated in these works such as the tiny transformer. In our work, we also seek to not only measure the impact of capacity but also to better understand why counter-intuitively higher levels of sparsity aid generalization. Finally, our experiments are extensive relative to \citep{Biljon2020OnOT,tapo-etal-2020-neural}, both in terms of number of languages and variety of training and evaluation conditions. Furthermore, we are the first to report human evaluation on the effects of pruning for MT.

\section{Future Work}
Our work introduces the term \textit{low-resource double-bind} and conducts extensive experiments to study the impact of pruning. In this setting, we are concerned with resource constraints present at deployment. An important area for further work is to explore a setting where resource constraints are present at both training and deployment time. For example, a consideration of the impact of pruning on pre-trained models, such as large multilingual MT models that are known to boost low-resource NMT quality~\citep{aharoni-etal-2019-massively,massiveWild}. Additionally, the minimal differences observed in our human evaluation of preferences open up a range of questions for deeper qualitative analysis of the resulting translations: Under which conditions do humans notice differences, and how do translations differ in style? There may be interesting connections to recent findings about output hallucinations occurring on memorized examples~\citep{curiousHallucinations}, or with respect to translation bias~\citep{koppel-ordan-2011-translationese}.

\section{Conclusion}\label{sec:conclusion}
We demonstrate the effectiveness of introducing sparsity when training NMT models for low-resourced languages. We show that small performance drops in extremely sparse regimes according to automatic metrics are not reflected in human-judged translation quality. Our extensive study of the impact of pruning on out-of-distribution generalization reveals that sparse models improve over dense models in a limited data regime.
Overall, these insights are promising for overcoming the low-resource double bind: Pruned models reduce resource requirements for deployment, and increase the robustness towards out-of-domain samples due to reduced memorization during training.

\section{Acknowledgements}\label{sec:acknowledgement}
We thank Colin Cherry, Rubungo Andre Niyongabo, Kelechi Ogueji and Trevor Gale for their invaluable feedback and comments on the paper draft. We also thank the anonymous reviewers for their time and comments on our paper. We thank CURe for providing compute credits, and the institutional support of Jonathan Caton and Natacha Mainville.
\newpage

\bibliography{anthology,custom}

\begin{thebibliography}{95}
\expandafter\ifx\csname natexlab\endcsname\relax\def\natexlab#1{#1}\fi

\bibitem[{ab~Tessera et~al.(2021)ab~Tessera, Hooker, and
  Rosman}]{tessera2021gradients}
Kale ab~Tessera, Sara Hooker, and Benjamin Rosman. 2021.
\newblock \href {http://arxiv.org/abs/2102.01670} {Keep the gradients flowing:
  Using gradient flow to study sparse network optimization}.

\bibitem[{Agi{\'c} and Vuli{\'c}(2019)}]{agic-vulic-2019-jw300}
{\v{Z}}eljko Agi{\'c} and Ivan Vuli{\'c}. 2019.
\newblock \href {https://doi.org/10.18653/v1/P19-1310} {{JW}300: A
  wide-coverage parallel corpus for low-resource languages}.
\newblock In \emph{Proceedings of the 57th Annual Meeting of the Association
  for Computational Linguistics}, pages 3204--3210, Florence, Italy.
  Association for Computational Linguistics.

\bibitem[{Aharoni et~al.(2019)Aharoni, Johnson, and
  Firat}]{aharoni-etal-2019-massively}
Roee Aharoni, Melvin Johnson, and Orhan Firat. 2019.
\newblock \href {https://doi.org/10.18653/v1/N19-1388} {Massively multilingual
  neural machine translation}.
\newblock In \emph{Proceedings of the 2019 Conference of the North {A}merican
  Chapter of the Association for Computational Linguistics: Human Language
  Technologies, Volume 1 (Long and Short Papers)}, pages 3874--3884,
  Minneapolis, Minnesota. Association for Computational Linguistics.

\bibitem[{Aji and Heafield(2020)}]{aji-heafield-2020-compressing}
Alham~Fikri Aji and Kenneth Heafield. 2020.
\newblock \href {https://doi.org/10.18653/v1/2020.ngt-1.4} {Compressing neural
  machine translation models with 4-bit precision}.
\newblock In \emph{Proceedings of the Fourth Workshop on Neural Generation and
  Translation}, pages 35--42, Online. Association for Computational
  Linguistics.

\bibitem[{Amodei et~al.(2018)Amodei, Hernandez, Sastry, Clark, Brockman, and
  Sutskever}]{2018Amodei}
Dario Amodei, Danny Hernandez, Girish Sastry, Jack Clark, Greg Brockman, and
  Ilya Sutskever. 2018.
\newblock \href {https://openai.com/blog/ai-and-compute/} {Ai and compute}.

\bibitem[{Anastasopoulos et~al.(2020)Anastasopoulos, Cattelan, Dou, Federico,
  Federmann, Genzel, Guzm{\'{a}}n, Hu, Hughes, Koehn, Lazar, Lewis, Neubig,
  Niu, {\"{O}}ktem, Paquin, Tang, and Tur}]{tico}
Antonios Anastasopoulos, Alessandro Cattelan, Zi{-}Yi Dou, Marcello Federico,
  Christian Federmann, Dmitriy Genzel, Francisco Guzm{\'{a}}n, Junjie Hu,
  Macduff Hughes, Philipp Koehn, Rosie Lazar, William Lewis, Graham Neubig,
  Mengmeng Niu, Alp {\"{O}}ktem, Eric Paquin, Grace Tang, and Sylwia Tur. 2020.
\newblock \href {http://arxiv.org/abs/2007.01788} {{TICO-19:} the translation
  initiative for covid-19}.
\newblock \emph{CoRR}, abs/2007.01788.

\bibitem[{Arivazhagan et~al.(2019)Arivazhagan, Bapna, Firat, Lepikhin, Johnson,
  Krikun, Chen, Cao, Foster, Cherry, Macherey, Chen, and Wu}]{massiveWild}
Naveen Arivazhagan, Ankur Bapna, Orhan Firat, Dmitry Lepikhin, Melvin Johnson,
  Maxim Krikun, Mia~Xu Chen, Yuan Cao, George~F. Foster, Colin Cherry, Wolfgang
  Macherey, Zhifeng Chen, and Yonghui Wu. 2019.
\newblock \href {http://arxiv.org/abs/1907.05019} {Massively multilingual
  neural machine translation in the wild: Findings and challenges}.
\newblock \emph{CoRR}, abs/1907.05019.

\bibitem[{Bai et~al.(2020)Bai, Zhang, Hou, Shang, Jin, Jiang, Liu, Lyu, and
  King}]{Bai2020BinaryBERTPT}
Haoli Bai, Wei Zhang, Lu~Hou, Lifeng Shang, Jing Jin, X.~Jiang, Qun Liu,
  Michael~R. Lyu, and Irwin King. 2020.
\newblock Binarybert: Pushing the limit of bert quantization.
\newblock \emph{ArXiv}, abs/2012.15701.

\bibitem[{Ba{\~n}{\'o}n et~al.(2020)Ba{\~n}{\'o}n, Chen, Haddow, Heafield,
  Hoang, Espl{\`a}-Gomis, Forcada, Kamran, Kirefu, Koehn, Ortiz~Rojas,
  Pla~Sempere, Ram{\'\i}rez-S{\'a}nchez, Sarr{\'\i}as, Strelec, Thompson,
  Waites, Wiggins, and Zaragoza}]{banon-etal-2020-paracrawl}
Marta Ba{\~n}{\'o}n, Pinzhen Chen, Barry Haddow, Kenneth Heafield, Hieu Hoang,
  Miquel Espl{\`a}-Gomis, Mikel~L. Forcada, Amir Kamran, Faheem Kirefu, Philipp
  Koehn, Sergio Ortiz~Rojas, Leopoldo Pla~Sempere, Gema
  Ram{\'\i}rez-S{\'a}nchez, Elsa Sarr{\'\i}as, Marek Strelec, Brian Thompson,
  William Waites, Dion Wiggins, and Jaume Zaragoza. 2020.
\newblock \href {https://doi.org/10.18653/v1/2020.acl-main.417} {{P}ara{C}rawl:
  Web-scale acquisition of parallel corpora}.
\newblock In \emph{Proceedings of the 58th Annual Meeting of the Association
  for Computational Linguistics}, pages 4555--4567, Online. Association for
  Computational Linguistics.

\bibitem[{Bender et~al.(2021)Bender, Gebru, McMillan-Major, and
  Shmitchell}]{bender_gebru_2021}
Emily~M. Bender, Timnit Gebru, Angelina McMillan-Major, and Shmargaret
  Shmitchell. 2021.
\newblock \href {https://doi.org/10.1145/3442188.3445922} {On the dangers of
  stochastic parrots: Can language models be too big?}
\newblock In \emph{Proceedings of the 2021 ACM Conference on Fairness,
  Accountability, and Transparency}, FAccT '21, page 610–623, New York, NY,
  USA. Association for Computing Machinery.

\bibitem[{Biljon et~al.(2020)Biljon, Pretorius, and Kreutzer}]{Biljon2020OnOT}
Elan~Van Biljon, Arnu Pretorius, and Julia Kreutzer. 2020.
\newblock On optimal transformer depth for low-resource language translation.
\newblock \emph{AfricaNLP Workshop}.

\bibitem[{Blalock et~al.(2020)Blalock, Ortiz, Frankle, and
  Guttag}]{blalock2020state}
Davis Blalock, Jose Javier~Gonzalez Ortiz, Jonathan Frankle, and John Guttag.
  2020.
\newblock What is the state of neural network pruning?
\newblock \emph{arXiv preprint arXiv:2003.03033}.

\bibitem[{Brown et~al.(2020)Brown, Bun, Feldman, Smith, and
  Talwar}]{brown2020memorization}
Gavin Brown, Mark Bun, Vitaly Feldman, Adam Smith, and Kunal Talwar. 2020.
\newblock \href {http://arxiv.org/abs/2012.06421} {When is memorization of
  irrelevant training data necessary for high-accuracy learning?}

\bibitem[{Card et~al.(2020)Card, Henderson, Khandelwal, Jia, Mahowald, and
  Jurafsky}]{card-etal-2020-little}
Dallas Card, Peter Henderson, Urvashi Khandelwal, Robin Jia, Kyle Mahowald, and
  Dan Jurafsky. 2020.
\newblock \href {https://doi.org/10.18653/v1/2020.emnlp-main.745} {With little
  power comes great responsibility}.
\newblock In \emph{Proceedings of the 2020 Conference on Empirical Methods in
  Natural Language Processing (EMNLP)}, pages 9263--9274, Online. Association
  for Computational Linguistics.

\bibitem[{Chen et~al.(2021)Chen, Cheng, Wang, Gan, Wang, and jing
  Liu}]{Chen2021EarlyBERTEB}
Xiao-Han Chen, Yu~Cheng, Shuohang Wang, Zhe Gan, Zhangyang Wang, and Jing jing
  Liu. 2021.
\newblock Earlybert: Efficient bert training via early-bird lottery tickets.
\newblock \emph{ArXiv}, abs/2101.00063.

\bibitem[{Chung et~al.(2020)Chung, Kim, Choi, Kwon, Jeon, Park, Kim, and
  Lee}]{chung-etal-2020-extremely}
Insoo Chung, Byeongwook Kim, Yoonjung Choi, Se~Jung Kwon, Yongkweon Jeon,
  Baeseong Park, Sangha Kim, and Dongsoo Lee. 2020.
\newblock \href {https://doi.org/10.18653/v1/2020.findings-emnlp.433}
  {Extremely low bit transformer quantization for on-device neural machine
  translation}.
\newblock In \emph{Findings of the Association for Computational Linguistics:
  EMNLP 2020}, pages 4812--4826, Online. Association for Computational
  Linguistics.

\bibitem[{Cieri et~al.(2016)Cieri, Maxwell, Strassel, and
  Tracey}]{cieri-etal-2016-selection}
Christopher Cieri, Mike Maxwell, Stephanie Strassel, and Jennifer Tracey. 2016.
\newblock \href {https://www.aclweb.org/anthology/L16-1720} {Selection criteria
  for low resource language programs}.
\newblock In \emph{Proceedings of the Tenth International Conference on
  Language Resources and Evaluation ({LREC}'16)}, pages 4543--4549,
  Portoro{\v{z}}, Slovenia. European Language Resources Association (ELRA).

\bibitem[{Collins and Kohli(2014)}]{memory-bounded-convnet}
Maxwell~D. Collins and Pushmeet Kohli. 2014.
\newblock \href {http://arxiv.org/abs/1412.1442} {Memory {B}ounded {D}eep
  {C}onvolutional {N}etworks}.
\newblock \emph{CoRR}, abs/1412.1442.

\bibitem[{{Courbariaux} et~al.(2014){Courbariaux}, {Bengio}, and
  {David}}]{2014Courbariaux_low_precision_multiplications}
Matthieu {Courbariaux}, Yoshua {Bengio}, and Jean-Pierre {David}. 2014.
\newblock \href {http://arxiv.org/abs/1412.7024} {{Training deep neural
  networks with low precision multiplications}}.
\newblock \emph{arXiv e-prints}, page arXiv:1412.7024.

\bibitem[{Cun et~al.(1990)Cun, Denker, and Solla}]{Cun90optimalbrain}
Yann~Le Cun, John~S. Denker, and Sara~A. Solla. 1990.
\newblock Optimal brain damage.
\newblock In \emph{Advances in Neural Information Processing Systems}, pages
  598--605. Morgan Kaufmann.

\bibitem[{Dossou and Emezue(2021)}]{OkwuGbe}
Bonaventure F.~P. Dossou and Chris~C. Emezue. 2021.
\newblock \href {https://arxiv.org/abs/2103.07762} {Okwugb{\'{e}}: End-to-end
  speech recognition for fon and igbo}.
\newblock \emph{AfricaNLP Workshop}.

\bibitem[{Duh et~al.(2020)Duh, McNamee, Post, and
  Thompson}]{duh-etal-2020-benchmarking}
Kevin Duh, Paul McNamee, Matt Post, and Brian Thompson. 2020.
\newblock \href {https://www.aclweb.org/anthology/2020.lrec-1.325}
  {Benchmarking neural and statistical machine translation on low-resource
  {A}frican languages}.
\newblock In \emph{Proceedings of the 12th Language Resources and Evaluation
  Conference}, pages 2667--2675, Marseille, France. European Language Resources
  Association.

\bibitem[{Edunov et~al.(2020)Edunov, Ott, Ranzato, and
  Auli}]{edunov-etal-2020-evaluation}
Sergey Edunov, Myle Ott, Marc{'}Aurelio Ranzato, and Michael Auli. 2020.
\newblock \href {https://doi.org/10.18653/v1/2020.acl-main.253} {On the
  evaluation of machine translation systems trained with back-translation}.
\newblock In \emph{Proceedings of the 58th Annual Meeting of the Association
  for Computational Linguistics}, pages 2836--2846, Online. Association for
  Computational Linguistics.

\bibitem[{Elsen et~al.(2019)Elsen, Dukhan, Gale, and Simonyan}]{elsen2019fast}
Erich Elsen, Marat Dukhan, Trevor Gale, and Karen Simonyan. 2019.
\newblock \href {http://arxiv.org/abs/1911.09723} {Fast sparse convnets}.

\bibitem[{Esteva et~al.(2017)Esteva, Kuprel, Novoa, Ko, M~Swetter, M~Blau, and
  Thrun}]{2017Andre}
Andre Esteva, Brett Kuprel, Roberto Novoa, Justin Ko, Susan M~Swetter, Helen
  M~Blau, and Sebastian Thrun. 2017.
\newblock \href {https://doi.org/10.1038/nature21056} {Dermatologist-level
  classification of skin cancer with deep neural networks}.
\newblock \emph{Nature}, 542.

\bibitem[{Evci et~al.(2019)Evci, Gale, Menick, Castro, and
  Elsen}]{evci2019rigging}
Utku Evci, Trevor Gale, Jacob Menick, Pablo~Samuel Castro, and Erich Elsen.
  2019.
\newblock \href {http://arxiv.org/abs/1911.11134} {Rigging the lottery: Making
  all tickets winners}.

\bibitem[{Fadaee et~al.(2017)Fadaee, Bisazza, and Monz}]{fadaee-etal-2017-data}
Marzieh Fadaee, Arianna Bisazza, and Christof Monz. 2017.
\newblock \href {https://doi.org/10.18653/v1/P17-2090} {Data augmentation for
  low-resource neural machine translation}.
\newblock In \emph{Proceedings of the 55th Annual Meeting of the Association
  for Computational Linguistics (Volume 2: Short Papers)}, pages 567--573,
  Vancouver, Canada. Association for Computational Linguistics.

\bibitem[{Feldman and Zhang(2020)}]{NEURIPS2020_1e14bfe2}
Vitaly Feldman and Chiyuan Zhang. 2020.
\newblock \href
  {https://proceedings.neurips.cc/paper/2020/file/1e14bfe2714193e7af5abc64ecbd6b46-Paper.pdf}
  {What neural networks memorize and why: Discovering the long tail via
  influence estimation}.
\newblock In \emph{Advances in Neural Information Processing Systems},
  volume~33, pages 2881--2891. Curran Associates, Inc.

\bibitem[{Gale et~al.(2019)Gale, Elsen, and Hooker}]{tgale_shooker_2019}
Trevor Gale, Erich Elsen, and Sara Hooker. 2019.
\newblock \href {http://arxiv.org/abs/1902.09574} {The state of sparsity in
  deep neural networks}.
\newblock \emph{CoRR}, abs/1902.09574.

\bibitem[{{Gale} et~al.(2019){Gale}, {Elsen}, and
  {Hooker}}]{2019arXiv190209574G}
Trevor {Gale}, Erich {Elsen}, and Sara {Hooker}. 2019.
\newblock \href {http://arxiv.org/abs/1902.09574} {{The State of Sparsity in
  Deep Neural Networks}}.
\newblock \emph{arXiv e-prints}, page arXiv:1902.09574.

\bibitem[{Gale et~al.(2020)Gale, Zaharia, Young, and Elsen}]{gale2020sparse}
Trevor Gale, Matei Zaharia, Cliff Young, and Erich Elsen. 2020.
\newblock \href {http://arxiv.org/abs/2006.10901} {Sparse gpu kernels for deep
  learning}.

\bibitem[{Goyal et~al.(2021)Goyal, Gao, Chaudhary, Chen, Wenzek, Ju, Krishnan,
  Ranzato, Guzm\'{a}n, and Fan}]{flores}
Naman Goyal, Cynthia Gao, Vishrav Chaudhary, Peng-Jen Chen, Guillaume Wenzek,
  Da~Ju, Sanjana Krishnan, Marc'Aurelio Ranzato, Francisco Guzm\'{a}n, and
  Angela Fan. 2021.
\newblock The flores-101 evaluation benchmark for low-resource and multilingual
  machine translation.

\bibitem[{Guo et~al.(2016)Guo, Yao, and Chen}]{dynamic-network-surgery}
Yiwen Guo, Anbang Yao, and Yurong Chen. 2016.
\newblock Dynamic {N}etwork {S}urgery for {E}fficient {DNN}s.
\newblock In \emph{NeurIPS}.

\bibitem[{Gupta et~al.(2015)Gupta, Agrawal, Gopalakrishnan, and
  Narayanan}]{2015_gupta}
Suyog Gupta, Ankur Agrawal, Kailash Gopalakrishnan, and Pritish Narayanan.
  2015.
\newblock \href {http://arxiv.org/abs/1502.02551} {Deep learning with limited
  numerical precision}.
\newblock \emph{CoRR}, abs/1502.02551.

\bibitem[{Han et~al.(2015)Han, Pool, Tran, and Dally}]{lwac}
Song Han, Jeff Pool, John Tran, and William~J. Dally. 2015.
\newblock {L}earning both {W}eights and {C}onnections for {E}fficient {N}eural
  {N}etwork.
\newblock In \emph{{NeurIPS}}, pages 1135--1143.

\bibitem[{Hassibi et~al.(1993{\natexlab{a}})Hassibi, Stork, and
  Wolff}]{1993optimalbrain}
B.~Hassibi, D.~G. Stork, and G.~J. Wolff. 1993{\natexlab{a}}.
\newblock \href {https://doi.org/10.1109/ICNN.1993.298572} {Optimal brain
  surgeon and general network pruning}.
\newblock In \emph{IEEE International Conference on Neural Networks}, pages
  293--299 vol.1.

\bibitem[{Hassibi et~al.(1993{\natexlab{b}})Hassibi, Stork, and
  Com}]{Hassibi93secondorder}
Babak Hassibi, David~G. Stork, and Stork Crc.~Ricoh. Com. 1993{\natexlab{b}}.
\newblock Second order derivatives for network pruning: Optimal brain surgeon.
\newblock In \emph{Advances in Neural Information Processing Systems 5}, pages
  164--171. Morgan Kaufmann.

\bibitem[{{Hinton} et~al.(2015){Hinton}, {Vinyals}, and {Dean}}]{2015hinton}
Geoffrey {Hinton}, Oriol {Vinyals}, and Jeff {Dean}. 2015.
\newblock \href {http://arxiv.org/abs/1503.02531} {{Distilling the Knowledge in
  a Neural Network}}.
\newblock \emph{arXiv e-prints}, page arXiv:1503.02531.

\bibitem[{Hooker et~al.(2020{\natexlab{a}})Hooker, Courville, Clark, Dauphin,
  and Frome}]{hooker2020compressed}
Sara Hooker, Aaron Courville, Gregory Clark, Yann Dauphin, and Andrea Frome.
  2020{\natexlab{a}}.
\newblock \href {http://arxiv.org/abs/1911.05248} {What do compressed deep
  neural networks forget?}

\bibitem[{Hooker et~al.(2020{\natexlab{b}})Hooker, Moorosi, Clark, Bengio, and
  Denton}]{hooker2020characterising}
Sara Hooker, Nyalleng Moorosi, Gregory Clark, Samy Bengio, and Emily Denton.
  2020{\natexlab{b}}.
\newblock \href {http://arxiv.org/abs/2010.03058} {Characterising bias in
  compressed models}.

\bibitem[{{Horowitz}(2014)}]{2014Horowitz}
M.~{Horowitz}. 2014.
\newblock 1.1 computing's energy problem (and what we can do about it).
\newblock In \emph{2014 IEEE International Solid-State Circuits Conference
  Digest of Technical Papers (ISSCC)}, pages 10--14.

\bibitem[{Hou et~al.(2020)Hou, Shang, Jiang, and Liu}]{Hou2020DynaBERTDB}
Lu~Hou, Lifeng Shang, X.~Jiang, and Qun Liu. 2020.
\newblock Dynabert: Dynamic bert with adaptive width and depth.
\newblock \emph{ArXiv}, abs/2004.04037.

\bibitem[{{Howard} et~al.(2017){Howard}, {Zhu}, {Chen}, {Kalenichenko}, {Wang},
  {Weyand}, {Andreetto}, and {Adam}}]{2017Howard}
A.~G. {Howard}, M.~{Zhu}, B.~{Chen}, D.~{Kalenichenko}, W.~{Wang}, T.~{Weyand},
  M.~{Andreetto}, and H.~{Adam}. 2017.
\newblock \href {http://arxiv.org/abs/1704.04861} {{MobileNets: Efficient
  Convolutional Neural Networks for Mobile Vision Applications}}.
\newblock \emph{ArXiv e-prints}.

\bibitem[{Hubara et~al.(2016)Hubara, Courbariaux, Soudry, El{-}Yaniv, and
  Bengio}]{Hubara2016_training_neural_networks_low_precision}
Itay Hubara, Matthieu Courbariaux, Daniel Soudry, Ran El{-}Yaniv, and Yoshua
  Bengio. 2016.
\newblock \href {http://arxiv.org/abs/1609.07061} {Quantized neural networks:
  Training neural networks with low precision weights and activations}.
\newblock \emph{CoRR}, abs/1609.07061.

\bibitem[{{Iandola} et~al.(2016){Iandola}, {Han}, {Moskewicz}, {Ashraf},
  {Dally}, and {Keutzer}}]{2016Squeezenet}
F.~N. {Iandola}, S.~{Han}, M.~W. {Moskewicz}, K.~{Ashraf}, W.~J. {Dally}, and
  K.~{Keutzer}. 2016.
\newblock \href {http://arxiv.org/abs/1602.07360} {{SqueezeNet: AlexNet-level
  accuracy with 50x fewer parameters and $<$0.5MB model size}}.
\newblock \emph{ArXiv e-prints}.

\bibitem[{Jacob et~al.(2018)Jacob, Kligys, Chen, Zhu, Tang, Howard, Adam, and
  Kalenichenko}]{Jacob_2018}
Benoit Jacob, Skirmantas Kligys, Bo~Chen, Menglong Zhu, Matthew Tang, Andrew
  Howard, Hartwig Adam, and Dmitry Kalenichenko. 2018.
\newblock \href {https://doi.org/10.1109/cvpr.2018.00286} {Quantization and
  training of neural networks for efficient integer-arithmetic-only inference}.
\newblock \emph{2018 IEEE/CVF Conference on Computer Vision and Pattern
  Recognition}.

\bibitem[{Joshi et~al.(2020)Joshi, Santy, Budhiraja, Bali, and
  Choudhury}]{joshi-etal-2020-state}
Pratik Joshi, Sebastin Santy, Amar Budhiraja, Kalika Bali, and Monojit
  Choudhury. 2020.
\newblock \href {https://doi.org/10.18653/v1/2020.acl-main.560} {The state and
  fate of linguistic diversity and inclusion in the {NLP} world}.
\newblock In \emph{Proceedings of the 58th Annual Meeting of the Association
  for Computational Linguistics}, pages 6282--6293, Online. Association for
  Computational Linguistics.

\bibitem[{Koppel and Ordan(2011)}]{koppel-ordan-2011-translationese}
Moshe Koppel and Noam Ordan. 2011.
\newblock \href {https://www.aclweb.org/anthology/P11-1132} {Translationese and
  its dialects}.
\newblock In \emph{Proceedings of the 49th Annual Meeting of the Association
  for Computational Linguistics: Human Language Technologies}, pages
  1318--1326, Portland, Oregon, USA. Association for Computational Linguistics.

\bibitem[{Kumar et~al.(2017)Kumar, Goyal, and Varma}]{kumar17}
Ashish Kumar, Saurabh Goyal, and Manik Varma. 2017.
\newblock \href {http://proceedings.mlr.press/v70/kumar17a.html}
  {Resource-efficient machine learning in 2 {KB} {RAM} for the internet of
  things}.
\newblock In \emph{Proceedings of the 34th International Conference on Machine
  Learning}, volume~70 of \emph{Proceedings of Machine Learning Research},
  pages 1935--1944, International Convention Centre, Sydney, Australia. PMLR.

\bibitem[{{Lane} and {Warden}(2018)}]{8364435}
N.~D. {Lane} and P.~{Warden}. 2018.
\newblock \href {https://doi.org/10.1109/MC.2018.2381129} {The deep (learning)
  transformation of mobile and embedded computing}.
\newblock \emph{Computer}, 51(5):12--16.

\bibitem[{Lazaridou et~al.(2021)Lazaridou, Kuncoro, Gribovskaya, Agrawal,
  Liska, Terzi, Gimenez, de~Masson~d'Autume, Ruder, Yogatama, Cao, Kocisky,
  Young, and Blunsom}]{lazaridou2021pitfalls}
Angeliki Lazaridou, Adhiguna Kuncoro, Elena Gribovskaya, Devang Agrawal, Adam
  Liska, Tayfun Terzi, Mai Gimenez, Cyprien de~Masson~d'Autume, Sebastian
  Ruder, Dani Yogatama, Kris Cao, Tomas Kocisky, Susannah Young, and Phil
  Blunsom. 2021.
\newblock \href {http://arxiv.org/abs/2102.01951} {Pitfalls of static language
  modelling}.

\bibitem[{Li et~al.(2020{\natexlab{a}})Li, Wang, Liu, Du, Xiao, Zhang, and
  Zhu}]{Li2020LearningLT}
Bei Li, Ziyang Wang, H.~Liu, Quan Du, Tong Xiao, Chunliang Zhang, and Jingbo
  Zhu. 2020{\natexlab{a}}.
\newblock Learning light-weight translation models from deep transformer.
\newblock \emph{ArXiv}, abs/2012.13866.

\bibitem[{Li et~al.(2020{\natexlab{b}})Li, Wallace, Shen, Lin, Keutzer, Klein,
  and Gonzalez}]{rethinking_model_size_li}
Zhuohan Li, Eric Wallace, Sheng Shen, Kevin Lin, Kurt Keutzer, Dan Klein, and
  Joseph~E. Gonzalez. 2020{\natexlab{b}}.
\newblock \href {https://arxiv.org/abs/2002.11794} {Train large, then compress:
  Rethinking model size for efficient training and inference of transformers}.
\newblock \emph{ICML}.

\bibitem[{Liebenwein et~al.(2021)Liebenwein, Baykal, Carter, Gifford, and
  Rus}]{liebenwein}
Lucas Liebenwein, Cenk Baykal, Brandon Carter, David Gifford, and Daniela Rus.
  2021.
\newblock \href {http://arxiv.org/abs/2103.03014} {Lost in pruning: The effects
  of pruning neural networks beyond test accuracy}.
\newblock \emph{CoRR}, abs/2103.03014.

\bibitem[{{Louizos} et~al.(2017){Louizos}, {Welling}, and
  {Kingma}}]{2017l0_reg}
C.~{Louizos}, M.~{Welling}, and D.~P. {Kingma}. 2017.
\newblock \href {http://arxiv.org/abs/1712.01312} {{Learning Sparse Neural
  Networks through $L\_0$ Regularization}}.
\newblock \emph{ArXiv e-prints}.

\bibitem[{Martinus and Abbott(2019)}]{afocus}
Laura Martinus and Jade~Z. Abbott. 2019.
\newblock \href {http://arxiv.org/abs/1906.05685} {A focus on neural machine
  translation for african languages}.
\newblock \emph{CoRR}, abs/1906.05685.

\bibitem[{Mocanu et~al.(2018)Mocanu, Mocanu, Stone, Nguyen, Gibescu, and
  Liotta}]{sparse-evolutionary-training}
Decebal~Constantin Mocanu, Elena Mocanu, Peter Stone, Phuong~H. Nguyen,
  Madeleine Gibescu, and Antonio Liotta. 2018.
\newblock Scalable {T}raining of {A}rtificial {N}eural {N}etworks with
  {A}daptive {S}parse {C}onnectivity {I}nspired by {N}etwork {S}cience.
\newblock \emph{Nature Communications}.

\bibitem[{Murray et~al.(2019)Murray, Kinnison, Nguyen, Scheirer, and
  Chiang}]{murray-etal-2019-auto}
Kenton Murray, Jeffery Kinnison, Toan~Q. Nguyen, Walter Scheirer, and David
  Chiang. 2019.
\newblock \href {https://doi.org/10.18653/v1/D19-5625} {Auto-sizing the
  transformer network: Improving speed, efficiency, and performance for
  low-resource machine translation}.
\newblock In \emph{Proceedings of the 3rd Workshop on Neural Generation and
  Translation}, pages 231--240, Hong Kong. Association for Computational
  Linguistics.

\bibitem[{{Narang} et~al.(2017){Narang}, {Elsen}, {Diamos}, and
  {Sengupta}}]{2017Narang}
Sharan {Narang}, Erich {Elsen}, Gregory {Diamos}, and Shubho {Sengupta}. 2017.
\newblock \href {http://arxiv.org/abs/1704.05119} {{Exploring Sparsity in
  Recurrent Neural Networks}}.
\newblock \emph{arXiv e-prints}, page arXiv:1704.05119.

\bibitem[{Narang et~al.(2017)Narang, Elsen, Diamos, and
  Sengupta}]{narang2017exploring}
Sharan Narang, Erich Elsen, Gregory Diamos, and Shubho Sengupta. 2017.
\newblock \href {http://arxiv.org/abs/1704.05119} {Exploring sparsity in
  recurrent neural networks}.

\bibitem[{Nekoto et~al.(2020)Nekoto, Marivate, Matsila, Fasubaa, Fagbohungbe,
  Akinola, Muhammad, Kabongo~Kabenamualu, Osei, Sackey, Niyongabo, Macharm,
  Ogayo, Ahia, Berhe, Adeyemi, Mokgesi-Selinga, Okegbemi, Martinus, Tajudeen,
  Degila, Ogueji, Siminyu, Kreutzer, Webster, Ali, Abbott, Orife, Ezeani,
  Dangana, Kamper, Elsahar, Duru, Kioko, Espoir, van Biljon, Whitenack,
  Onyefuluchi, Emezue, Dossou, Sibanda, Bassey, Olabiyi, Ramkilowan, {\"O}ktem,
  Akinfaderin, and Bashir}]{nekoto-etal-2020-participatory}
Wilhelmina Nekoto, Vukosi Marivate, Tshinondiwa Matsila, Timi Fasubaa, Taiwo
  Fagbohungbe, Solomon~Oluwole Akinola, Shamsuddeen Muhammad, Salomon
  Kabongo~Kabenamualu, Salomey Osei, Freshia Sackey, Rubungo~Andre Niyongabo,
  Ricky Macharm, Perez Ogayo, Orevaoghene Ahia, Musie~Meressa Berhe, Mofetoluwa
  Adeyemi, Masabata Mokgesi-Selinga, Lawrence Okegbemi, Laura Martinus,
  Kolawole Tajudeen, Kevin Degila, Kelechi Ogueji, Kathleen Siminyu, Julia
  Kreutzer, Jason Webster, Jamiil~Toure Ali, Jade Abbott, Iroro Orife, Ignatius
  Ezeani, Idris~Abdulkadir Dangana, Herman Kamper, Hady Elsahar, Goodness Duru,
  Ghollah Kioko, Murhabazi Espoir, Elan van Biljon, Daniel Whitenack,
  Christopher Onyefuluchi, Chris~Chinenye Emezue, Bonaventure F.~P. Dossou,
  Blessing Sibanda, Blessing Bassey, Ayodele Olabiyi, Arshath Ramkilowan, Alp
  {\"O}ktem, Adewale Akinfaderin, and Abdallah Bashir. 2020.
\newblock \href {https://doi.org/10.18653/v1/2020.findings-emnlp.195}
  {Participatory research for low-resourced machine translation: A case study
  in {A}frican languages}.
\newblock In \emph{Findings of the Association for Computational Linguistics:
  EMNLP 2020}, pages 2144--2160, Online. Association for Computational
  Linguistics.

\bibitem[{{\"{O}}ktem et~al.(2021){\"{O}}ktem, DeLuca, Bashizi, Paquin, and
  Tang}]{congolese}
Alp {\"{O}}ktem, Eric DeLuca, Rodrigue Bashizi, Eric Paquin, and Grace Tang.
  2021.
\newblock \href {https://arxiv.org/abs/2103.10734} {Congolese swahili machine
  translation for humanitarian response}.
\newblock \emph{AfricaNLP Workshop}.

\bibitem[{{\"{O}}ktem et~al.(2020){\"{O}}ktem, Plitt, and Tang}]{tigrinya}
Alp {\"{O}}ktem, Mirko Plitt, and Grace Tang. 2020.
\newblock \href {https://arxiv.org/abs/2003.11523} {Tigrinya neural machine
  translation with transfer learning for humanitarian response}.
\newblock \emph{AfricaNLP Workshop}.

\bibitem[{Orife(2018)}]{orifeDiacritics}
Iroro Orife. 2018.
\newblock \href {http://arxiv.org/abs/1804.00832} {Attentive
  sequence-to-sequence learning for diacritic restoration of yor{\`{u}}b{\'{a}}
  language text}.
\newblock \emph{CoRR}, abs/1804.00832.

\bibitem[{Oughton(2021)}]{oughton2021}
Edward~J. Oughton. 2021.
\newblock \href {http://arxiv.org/abs/2102.03561} {Policy options for digital
  infrastructure strategies: {A} simulation model for broadband universal
  service in africa}.
\newblock \emph{CoRR}, abs/2102.03561.

\bibitem[{Papineni et~al.(2002)Papineni, Roukos, Ward, and Zhu}]{Papineni2002}
Kishore Papineni, Salim Roukos, Todd Ward, and Wei-Jing Zhu. 2002.
\newblock \href {https://doi.org/10.3115/1073083.1073135} {Bleu: A method for
  automatic evaluation of machine translation}.
\newblock In \emph{Proceedings of the 40th Annual Meeting on Association for
  Computational Linguistics}, ACL '02, page 311–318, USA. Association for
  Computational Linguistics.

\bibitem[{Patterson et~al.(2021)Patterson, Gonzalez, Le, Liang, Munguia,
  Rothchild, So, Texier, and Dean}]{patterson2021carbon}
David Patterson, Joseph Gonzalez, Quoc Le, Chen Liang, Lluis-Miquel Munguia,
  Daniel Rothchild, David So, Maud Texier, and Jeff Dean. 2021.
\newblock \href {http://arxiv.org/abs/2104.10350} {Carbon emissions and large
  neural network training}.

\bibitem[{Post(2018)}]{post-2018-call}
Matt Post. 2018.
\newblock \href {https://doi.org/10.18653/v1/W18-6319} {A call for clarity in
  reporting {BLEU} scores}.
\newblock In \emph{Proceedings of the Third Conference on Machine Translation:
  Research Papers}, pages 186--191, Brussels, Belgium. Association for
  Computational Linguistics.

\bibitem[{Quinn and Ballesteros(2018)}]{quinn-ballesteros-2018-pieces}
Jerry Quinn and Miguel Ballesteros. 2018.
\newblock \href {https://doi.org/10.18653/v1/N18-3014} {Pieces of eight: 8-bit
  neural machine translation}.
\newblock In \emph{Proceedings of the 2018 Conference of the North {A}merican
  Chapter of the Association for Computational Linguistics: Human Language
  Technologies, Volume 3 (Industry Papers)}, pages 114--120, New Orleans -
  Louisiana. Association for Computational Linguistics.

\bibitem[{Raunak et~al.(2020)Raunak, Dalmia, Gupta, and
  Metze}]{raunak-etal-2020-long}
Vikas Raunak, Siddharth Dalmia, Vivek Gupta, and Florian Metze. 2020.
\newblock \href {https://doi.org/10.18653/v1/2020.findings-emnlp.276} {On
  long-tailed phenomena in neural machine translation}.
\newblock In \emph{Findings of the Association for Computational Linguistics:
  EMNLP 2020}, pages 3088--3095, Online. Association for Computational
  Linguistics.

\bibitem[{Raunak et~al.(2021)Raunak, Menezes, and
  Junczys{-}Dowmunt}]{curiousHallucinations}
Vikas Raunak, Arul Menezes, and Marcin Junczys{-}Dowmunt. 2021.
\newblock \href {http://arxiv.org/abs/2104.06683} {The curious case of
  hallucinations in neural machine translation}.
\newblock \emph{CoRR}, abs/2104.06683.

\bibitem[{Reed(1993)}]{248452}
R.~Reed. 1993.
\newblock \href {https://doi.org/10.1109/72.248452} {Pruning algorithms-a
  survey}.
\newblock \emph{IEEE Transactions on Neural Networks}, 4(5):740--747.

\bibitem[{Samala et~al.(2018)Samala, Chan, Hadjiiski, Helvie, Richter, and
  Cha}]{Samala_2018}
Ravi~K Samala, Heang-Ping Chan, Lubomir~M Hadjiiski, Mark~A Helvie, Caleb
  Richter, and Kenny Cha. 2018.
\newblock \href {https://doi.org/10.1088/1361-6560/aabb5b} {Evolutionary
  pruning of transfer learned deep convolutional neural network for breast
  cancer diagnosis in digital breast tomosynthesis}.
\newblock \emph{Physics in Medicine {\&} Biology}, 63(9):095005.

\bibitem[{Sanh et~al.(2020)Sanh, Wolf, and Rush}]{sanh2020movement}
Victor Sanh, Thomas Wolf, and Alexander~M. Rush. 2020.
\newblock \href {http://arxiv.org/abs/2005.07683} {Movement pruning: Adaptive
  sparsity by fine-tuning}.

\bibitem[{{See} et~al.(2016){See}, {Luong}, and {Manning}}]{2016abigail}
Abigail {See}, Minh-Thang {Luong}, and Christopher~D. {Manning}. 2016.
\newblock \href {http://arxiv.org/abs/1606.09274} {{Compression of Neural
  Machine Translation Models via Pruning}}.
\newblock \emph{arXiv e-prints}, page arXiv:1606.09274.

\bibitem[{See et~al.(2016)See, Luong, and Manning}]{see-etal-2016-compression}
Abigail See, Minh-Thang Luong, and Christopher~D. Manning. 2016.
\newblock \href {https://doi.org/10.18653/v1/K16-1029} {Compression of neural
  machine translation models via pruning}.
\newblock In \emph{Proceedings of The 20th {SIGNLL} Conference on Computational
  Natural Language Learning}, pages 291--301, Berlin, Germany. Association for
  Computational Linguistics.

\bibitem[{Sennrich et~al.(2016)Sennrich, Haddow, and
  Birch}]{sennrich-etal-2016-neural}
Rico Sennrich, Barry Haddow, and Alexandra Birch. 2016.
\newblock \href {https://doi.org/10.18653/v1/P16-1162} {Neural machine
  translation of rare words with subword units}.
\newblock In \emph{Proceedings of the 54th Annual Meeting of the Association
  for Computational Linguistics (Volume 1: Long Papers)}, pages 1715--1725,
  Berlin, Germany. Association for Computational Linguistics.

\bibitem[{Sennrich and Zhang(2019)}]{sennrich-zhang-2019-revisiting}
Rico Sennrich and Biao Zhang. 2019.
\newblock \href {https://doi.org/10.18653/v1/P19-1021} {Revisiting low-resource
  neural machine translation: A case study}.
\newblock In \emph{Proceedings of the 57th Annual Meeting of the Association
  for Computational Linguistics}, pages 211--221, Florence, Italy. Association
  for Computational Linguistics.

\bibitem[{S{\o}gaard et~al.(2021)S{\o}gaard, Ebert, Bastings, and
  Filippova}]{sogaard-etal-2021-need}
Anders S{\o}gaard, Sebastian Ebert, Jasmijn Bastings, and Katja Filippova.
  2021.
\newblock \href {https://www.aclweb.org/anthology/2021.eacl-main.156} {We need
  to talk about random splits}.
\newblock In \emph{Proceedings of the 16th Conference of the European Chapter
  of the Association for Computational Linguistics: Main Volume}, pages
  1823--1832, Online. Association for Computational Linguistics.

\bibitem[{Strubell et~al.(2019)Strubell, Ganesh, and
  McCallum}]{strubelL2019energy}
Emma Strubell, Ananya Ganesh, and Andrew McCallum. 2019.
\newblock \href {http://arxiv.org/abs/1906.02243} {Energy and policy
  considerations for deep learning in nlp}.

\bibitem[{Ström(1997)}]{Strom97sparseconnection}
Nikko Ström. 1997.
\newblock Sparse connection and pruning in large dynamic artificial neural
  networks.

\bibitem[{Sun et~al.(2020)Sun, Qin, Zhang, Liu, Chen, and
  Xie}]{sun_computation_sparse}
Fei Sun, Minghai Qin, Tianyun Zhang, Liu Liu, Yen-Kuang Chen, and Yuan Xie.
  2020.
\newblock Computation on sparse neural networks and its implications for future
  hardware: Invited.
\newblock In \emph{Proceedings of the 57th ACM/EDAC/IEEE Design Automation
  Conference}, DAC '20. IEEE Press.

\bibitem[{Tapo et~al.(2020)Tapo, Coulibaly, Diarra, Homan, Kreutzer, Luger,
  Nagashima, Zampieri, and Leventhal}]{tapo-etal-2020-neural}
Allahsera~Auguste Tapo, Bakary Coulibaly, S{\'e}bastien Diarra, Christopher
  Homan, Julia Kreutzer, Sarah Luger, Arthur Nagashima, Marcos Zampieri, and
  Michael Leventhal. 2020.
\newblock \href {https://www.aclweb.org/anthology/2020.loresmt-1.3} {Neural
  machine translation for extremely low-resource {A}frican languages: A case
  study on {B}ambara}.
\newblock In \emph{Proceedings of the 3rd Workshop on Technologies for MT of
  Low Resource Languages}, pages 23--32, Suzhou, China. Association for
  Computational Linguistics.

\bibitem[{Tapo et~al.(2021)Tapo, Leventhal, Luger, Homan, and
  Zampieri}]{Tapo2021}
Allahsera~Auguste Tapo, Michael Leventhal, Sarah Luger, Christopher~M. Homan,
  and Marcos Zampieri. 2021.
\newblock \href {https://arxiv.org/abs/2104.00041} {Domain-specific {MT} for
  low-resource languages: The case of bambara-french}.
\newblock \emph{AfricaNLP Workshop}.

\bibitem[{{Thompson} et~al.(2020){Thompson}, {Greenewald}, {Lee}, and
  {Manso}}]{2020arXiv200705558T}
Neil~C. {Thompson}, Kristjan {Greenewald}, Keeheon {Lee}, and Gabriel~F.
  {Manso}. 2020.
\newblock \href {http://arxiv.org/abs/2007.05558} {{The Computational Limits of
  Deep Learning}}.
\newblock \emph{arXiv e-prints}, page arXiv:2007.05558.

\bibitem[{Vaswani et~al.(2018)Vaswani, Bengio, Brevdo, Chollet, Gomez, Gouws,
  Jones, Kaiser, Kalchbrenner, Parmar, Sepassi, Shazeer, and
  Uszkoreit}]{tensor2tensor}
Ashish Vaswani, Samy Bengio, Eugene Brevdo, Francois Chollet, Aidan~N. Gomez,
  Stephan Gouws, Llion Jones, \L{}ukasz Kaiser, Nal Kalchbrenner, Niki Parmar,
  Ryan Sepassi, Noam Shazeer, and Jakob Uszkoreit. 2018.
\newblock \href {http://arxiv.org/abs/1803.07416} {Tensor2tensor for {N}eural
  {M}achine {T}ranslation}.
\newblock \emph{CoRR}, abs/1803.07416.

\bibitem[{Vaswani et~al.(2017)Vaswani, Shazeer, Parmar, Uszkoreit, Jones,
  Gomez, Kaiser, and Polosukhin}]{vaswani2017attention}
Ashish Vaswani, Noam Shazeer, Niki Parmar, Jakob Uszkoreit, Llion Jones,
  Aidan~N Gomez, {\L}ukasz Kaiser, and Illia Polosukhin. 2017.
\newblock Attention is all you need.
\newblock In \emph{Advances in neural information processing systems}, pages
  5998--6008.

\bibitem[{Warden and Situnayake(2019)}]{warden2019tinyml}
P.~Warden and D.~Situnayake. 2019.
\newblock \href {https://books.google.com/books?id=sB3mxQEACAAJ} {\emph{TinyML:
  Machine Learning with TensorFlow Lite on Arduino and Ultra-Low-Power
  Microcontrollers}}.
\newblock O'Reilly Media, Incorporated.

\bibitem[{{Wen} et~al.(2016){Wen}, {Wu}, {Wang}, {Chen}, and
  {Li}}]{2016learnedSparsity}
W.~{Wen}, C.~{Wu}, Y.~{Wang}, Y.~{Chen}, and H.~{Li}. 2016.
\newblock \href {http://arxiv.org/abs/1608.03665} {{Learning Structured
  Sparsity in Deep Neural Networks}}.
\newblock \emph{ArXiv e-prints}.

\bibitem[{Xu et~al.(2020)Xu, Hu, Jiang, Feng, Wang, Huang, Ju, Xiao, and
  Zhu}]{xu-etal-2020-dynamic}
Chen Xu, Bojie Hu, Yufan Jiang, Kai Feng, Zeyang Wang, Shen Huang, Qi~Ju, Tong
  Xiao, and Jingbo Zhu. 2020.
\newblock \href {https://doi.org/10.18653/v1/2020.coling-main.352} {Dynamic
  curriculum learning for low-resource neural machine translation}.
\newblock In \emph{Proceedings of the 28th International Conference on
  Computational Linguistics}, pages 3977--3989, Barcelona, Spain (Online).
  International Committee on Computational Linguistics.

\bibitem[{Zhang et~al.(2017)Zhang, Chen, Wang, Xu, and
  Xu}]{zhang-etal-2017-towards}
Xiaowei Zhang, Wei Chen, Feng Wang, Shuang Xu, and Bo~Xu. 2017.
\newblock \href {https://doi.org/10.18653/v1/D17-1154} {Towards compact and
  fast neural machine translation using a combined method}.
\newblock In \emph{Proceedings of the 2017 Conference on Empirical Methods in
  Natural Language Processing}, pages 1475--1481, Copenhagen, Denmark.
  Association for Computational Linguistics.

\bibitem[{Zhu et~al.(2019)Zhu, Zhang, Gu, and Xie}]{sparse_tensor_core}
Maohua Zhu, Tao Zhang, Zhenyu Gu, and Yuan Xie. 2019.
\newblock \href {https://doi.org/10.1145/3352460.3358269} {Sparse tensor core:
  Algorithm and hardware co-design for vector-wise sparse neural networks on
  modern gpus}.
\newblock In \emph{Proceedings of the 52nd Annual IEEE/ACM International
  Symposium on Microarchitecture}, MICRO '52, page 359–371, New York, NY,
  USA. Association for Computing Machinery.

\bibitem[{Zhu and Gupta(2017)}]{to-prune-or-not}
Michael Zhu and Suyog Gupta. 2017.
\newblock \href {http://arxiv.org/abs/1710.01878} {To prune, or not to prune:
  exploring the efficacy of pruning for model compression}.
\newblock \emph{CoRR}, abs/1710.01878.

\bibitem[{Zipf(1999)}]{zipf1999psycho}
G.K. Zipf. 1999.
\newblock \href {https://books.google.com/books?id=w1Z4Aq-5sWMC} {\emph{The
  Psycho-Biology of Language: An Introduction to Dynamic Philology}}.
\newblock Cognitive psychology]. Routledge.

\bibitem[{Zoph et~al.(2016)Zoph, Yuret, May, and
  Knight}]{zoph-etal-2016-transfer}
Barret Zoph, Deniz Yuret, Jonathan May, and Kevin Knight. 2016.
\newblock \href {https://doi.org/10.18653/v1/D16-1163} {Transfer learning for
  low-resource neural machine translation}.
\newblock In \emph{Proceedings of the 2016 Conference on Empirical Methods in
  Natural Language Processing}, pages 1568--1575, Austin, Texas. Association
  for Computational Linguistics.

\end{thebibliography}
\bibliographystyle{acl_natbib}

\appendix
\clearpage

\section{Training}
\label{sec:appendix}

\subsection{Overview of Compression}\label{sec:compression}

Popular model compression research directions includes reducing the precision or bit size per model weight (quantization) \citep{Jacob_2018, 2014Courbariaux_low_precision_multiplications, Hubara2016_training_neural_networks_low_precision, 2015_gupta}, efforts to start with a network that is more compact with fewer parameters, layers or computations (architecture design) \citep{2017Howard, 2016Squeezenet, kumar17}, student networks with fewer parameters that learn from a larger teacher model (model distillation) \citep{2015hinton} and finally pruning by setting a subset of weights or filters to zero \citep{2017l0_reg, 2016learnedSparsity, Cun90optimalbrain, 1993optimalbrain, Strom97sparseconnection, Hassibi93secondorder, 2016abigail, 2017Narang}. Often, a combination of compression methods might be applied. For example, pruning might be combined with other efficiency-improving methods, e.g. quantization or faster search algorithms. Quantization can be used to speed up inference and relax hardware requirements, as has been shown for e.g.,  8-bit ~\citep{quinn-ballesteros-2018-pieces},  
 4-bit~\citep{aji-heafield-2020-compressing} and recently also below 3-bit quantization~\citep{chung-etal-2020-extremely} of NMT models. In the wider NLP space, there has been interest in evaluating the trade-offs of different compression techniques for downstream finetuning. ~\citet{sanh2020movement} propose the use of movement pruning, a simple, deterministic first-order weight pruning method that is more adaptive to pre-trained model fine-tuning.

Magnitude-based weight pruning schemes use the magnitude of each weight as a proxy for its importance to model quality, and remove the least important weights according to some sparsification schedule over the course of training. Many variants have been proposed \cite{memory-bounded-convnet, lwac, dynamic-network-surgery, to-prune-or-not}, which can be distinguished by differences in the criteria used to remove weights, when weights are removed and whether weights that have been pruned can still receive gradient updates after being removed.

\citet{lwac} use iterative magnitude pruning and re-training to progressively sparsify a model. The target model is first trained to convergence, after which a portion of weights are removed and the model is re-trained with these weights fixed to zero. This process is repeated until the target sparsity is achieved. \citet{dynamic-network-surgery} improve on this approach by allowing masked weights to still receive gradient updates, enabling the network to recover from incorrect pruning decisions during optimization. They achieve higher compression rates and interleave pruning steps with gradient update steps to avoid expensive re-training. \citet{to-prune-or-not} similarly allow gradient updates to masked weights, and make use of a gradual sparsification schedule with sorting-based weight thresholding to maintain accuracy while achieving a user specified level of sparsification. 

Magnitude pruning can easily be adapted to induce block or activation level sparsity by removing groups of weights based on their p-norm, average, max, or other statistics. Variants have also been proposed that maintain a constant level of sparsity during optimization to enable accelerated training \cite{sparse-evolutionary-training}.

\subsection{Architecture Size}
Table~\ref{tab:sizes} compares the sizes for base and tiny transformers. We do a model size ablation, comparing two model types (sizes in Appendix Table \ref{tab:sizes}). Table \ref{tab:parameter_count} displays the results showing that the \texttt{tiny} transformer gives a lower BLEU score than the \texttt{base} transformer even with extensive hyperparameter tuning. Hence we use the sparse transformer \texttt{base} model for all our preferred experiments.

\subsection{Training Hyperparameters}\label{sec:hyper-training}
We train the transformer with the hyper-parameters and optimizer settings described in~\citep{vaswani2017attention}. We use a batch size of 2048, and train on a Google Cloud TPU v2-8 with a default learning rate of 0.2, and learning rate warm-up steps of 8000. Regularization is introduced with dropout and label smoothing with rates 0.1. We are interested in a setting where resource constraints are present at deployment time, and do not assume constraints present at training. Our code is publicly available at \url{https://github.com/orevaahia/mc4lrnmt}.
In our experiments on the \texttt{Full} dataset, models for Yoruba and Igbo and German are trained for a total number of 100k steps while the Hausa models are trained for 60k steps. For our experiments in a data limited regime, we  train for a total of 60k steps across all languages. 

\subsection{Pruning Hyperparameters}\label{sec:hyper}
We perform a limited hyperparameter search with manual tuning to determine the best pruning start time, end time and recovery interval and select the best models based on the BLEU performance, we notice that when we train on the \texttt{Full} data, we see an average difference of 0.7 if we introduce pruning early and stop pruning close to the end of training on all sparsity levels we train for. In our case, we begin pruning on the 2000th train step and end pruning on the 80000th step for German, Igbo and Yoruba and begin pruning on the 2000th train step and end pruning on the 40000th for Hausa. This is line with the results reported by \citet{2019arXiv190209574G} when evaluating different compression techniques on high resourced languages. Training on \texttt{limited} data however shows slightly different results. In most cases, we see also see an average increase of 0.7 when we start to prune at nearly a quarter of the total train steps and stop pruning 20,000 steps before training ends. For all experiments on the \texttt{Full} data, the frequency of pruning is every \SI{2000} steps, however for \texttt{limited} data experiments it is either 1000 or 2000 steps. Full tuning results are in Table~\ref{tab:tuning_hparams_samp}.

\begin{table}[t]
\centering
\begin{tabular}{lc c} 
\toprule
\textbf{Hyperparameters} & \textbf{Base}  & \textbf{Tiny}  \\
\midrule
Transformer Layers & 6 & 2 \\
Hidden Size & 512 & 128 \\
Attention Heads & 8 & 4 \\
Filter Size & 2048 & 512 \\ 
Optimizer & adafactor & adafactor \\
Max Length  & 64 & 64 \\
\bottomrule
\end{tabular}
\caption{Hyperparameters for transformer variants.}
\label{tab:sizes}
\end{table}

\section{Human Evaluation Details}~\label{app:human}
Test set quality is evaluated independently, translations for rating were randomly selected from each test set that yielded different translations from the two models. The ratio of identical translations that are withheld from this rating lies around 27\% for all languages.

The absolute ratings allow us to draw conclusions about the absolute quality, and the presentation in pairs encourages the rater to consider differences between both translations for their rating. We gather three independent ratings for translations into German, and one rating for translations into Yoruba, Hausa and Igbo. Ratings from multiple raters are aggregated by using the median score. 

Table~\ref{tab:ratings_global} reports absolute scores as well as wins/losses of sentence-level comparisons.

\begin{table}[t]
    \centering
    \resizebox{0.9\columnwidth}{!}{
    \begin{tabular}{l|cc|ccc}
    \toprule 
        & \multicolumn{2}{c|}{\textbf{Absolute Ratings}} & \multicolumn{3}{c}{\textbf{Relative Wins [\%]}} \\ 
    Sparsity    & 0 & 90 & 0 & Neither & 90 \\
        \midrule
        \texttt{de} & 4.17 & 4.06 & 31 & 46 & 23 \\
        \texttt{yo} & 3.66 & 3.66 & 26 & 48 & 26 \\ 
        \texttt{ig} & 3.87 & 3.96 & 31 & 31 & 37 \\ 
        \texttt{ha} & 4.77 & 4.85 & 29 & 38 & 34 \\ 
    \bottomrule
    \end{tabular}%
    }
    \caption{Results of the human evaluation study of 500 \texttt{Global} test set translations comparing \textbf{dense} (0) and \textbf{90\%-sparse} (90) models (\texttt{Limited}). 
    Absolute ratings are averaged across sentences.}
    \label{tab:ratings_global}
    	\vspace{-0.2cm}
\end{table}

\begin{table}[t]
    \centering
    \resizebox{0.9\columnwidth}{!}{
    \begin{tabular}{l|cc|ccc}
    \toprule 
         & \multicolumn{2}{c|}{\textbf{Absolute Ratings}} & \multicolumn{3}{c}{\textbf{Relative Wins [\%]}} \\  
    Sparsity    & 0 & 90 & 0 &  Neither & 90 \\
        \midrule
        \texttt{de} & 3.80 & 3.66 & 29 & 50 & 21 \\
        \texttt{yo} & 3.51 & 3.51 & 19 & 61 & 19 \\ 
        \texttt{ig} & 3.81 & 3.85 & 35 & 28 & 37 \\ 
        \texttt{ha} & 4.53 & 4.53 & 26 & 47 & 27\\ 
    \bottomrule
    \end{tabular}%
    }
    \caption{Results of the human evaluation study of 500 \texttt{Random} test set translations comparing \textbf{dense} (0) and \textbf{90\%-sparse} (90) models (\texttt{Limited}). 
    Absolute ratings are averaged across sentences.}
    \label{tab:ratings_global}
    	\vspace{-0.2cm}
\end{table}

\section{Distribution Shift Evaluation Results}
\begin{figure*}
	\centering
	\vskip 0.15in
\begin{small}
\begin{sc}
	\begin{subfigure}{0.33\linewidth}
		\centering	\includegraphics[width=0.99\linewidth]{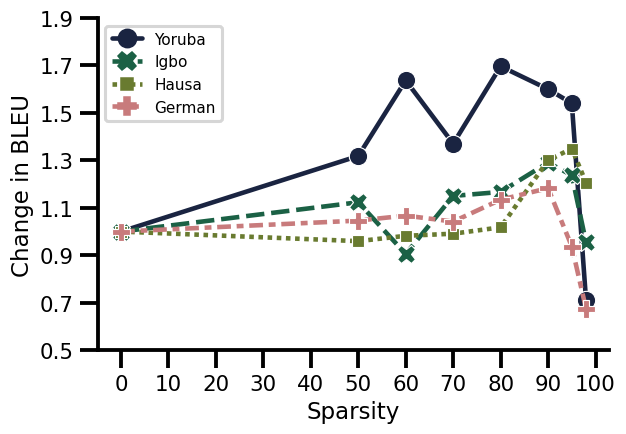}
    		\caption{Gnome}
	\end{subfigure}
		\begin{subfigure}{0.33\linewidth}
		\centering
			
    	\includegraphics[width=0.99\linewidth]{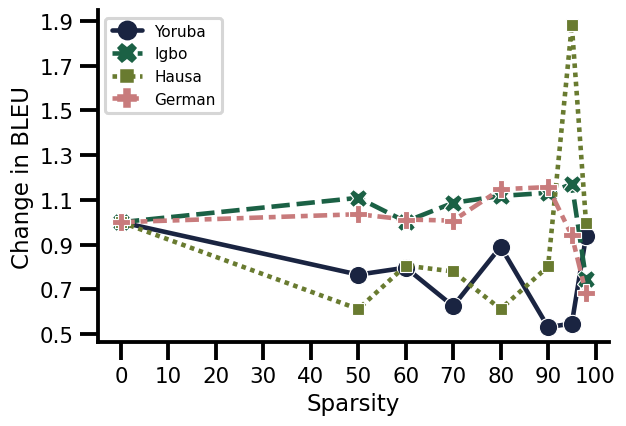}
    		\caption{Ubuntu)}
	\end{subfigure}
	\begin{subfigure}{0.33\linewidth}
		\centering
			
    	\includegraphics[width=0.99\linewidth]{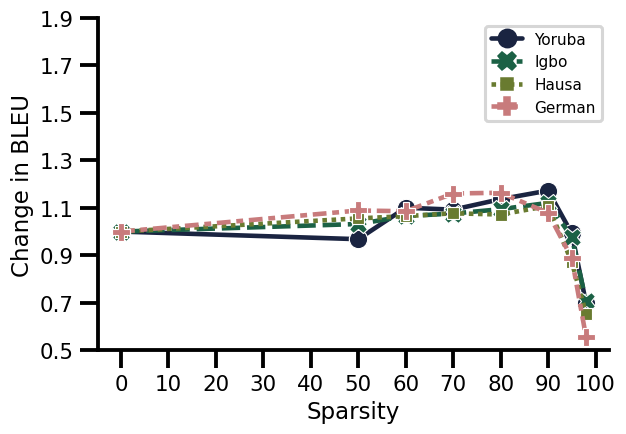}
    		\caption{Flores)}
	\end{subfigure}
\\
	\begin{subfigure}{0.35\linewidth}
		\centering
    	\includegraphics[width=0.99\linewidth]{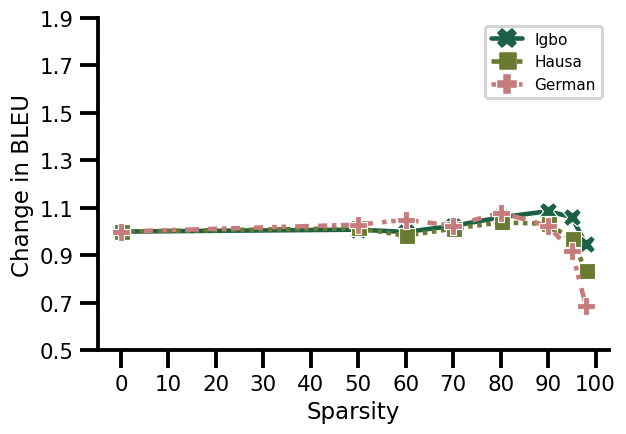}
    		\caption{ParaCrawl}
	\end{subfigure}	
	\begin{subfigure}{0.35\linewidth}
		\centering	\includegraphics[width=0.99\linewidth]{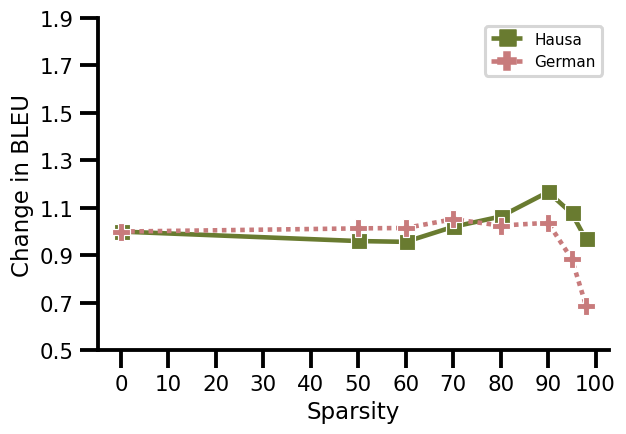}
\caption{Tanzil}
	\end{subfigure}	
	\end{sc}
	\end{small}
	\caption{
	   Relative change in BLEU on distribution shift datasets across languages and sparsity levels.
	}
	\label{fig:relative_degradation_distribution2}
	\vspace{-0.5cm}
\end{figure*}

\begin{figure*}
	\centering
	\vskip 0.15in
\begin{small}
\begin{sc}
	\begin{subfigure}{0.40\linewidth}
		\centering	\includegraphics[width=0.99\linewidth]{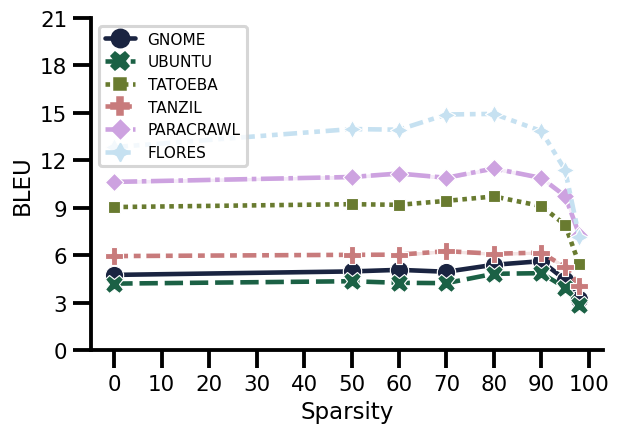}
    		\caption{\texttt{Limited}}
	\end{subfigure}
		\begin{subfigure}{0.40\linewidth}
		\centering
			
    	\includegraphics[width=0.99\linewidth]{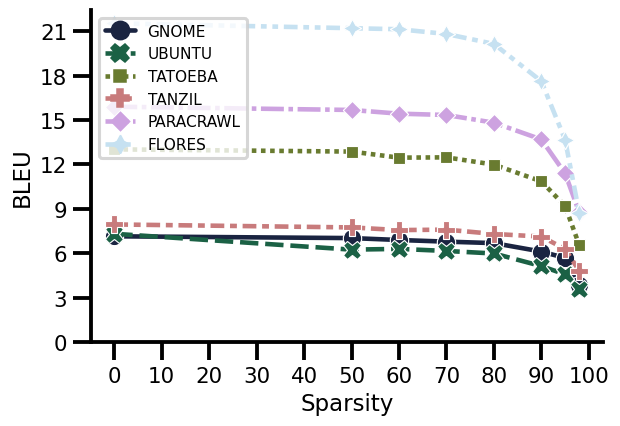}
    		\caption{\texttt{Full}}
	\end{subfigure}

	\end{sc}
	\end{small}
	\caption{
	   Absolute BLEU performance on out-of-distribution data for German across all sparsity levels. 
	}
	\label{fig:sensitivity_german}
	\vspace{-0.5cm}
\end{figure*}

We provide multiple views on translation quality under distribution shift for all languages in Figures~\ref{fig:sensitivity} and~\ref{fig:relative_degradation_distribution2}. To ease comparison, we summarize all results for German in Figure~\ref{fig:sensitivity_german} since it is present in all the datasets we consider. Relative performance degradation is measured by dividing the BLEU \citep{Papineni2002} from the sparse models by that of the dense model.

\section{Results for Full vs Limited Training Regime}
Figure~\ref{fig:degradation} shows the relative and absolute differences in BLEU caused by pruning in both \texttt{Limited} and \texttt{Full} models. We see minimal changes in BLEU between both training regimes and conclude that the ranking of absolute BLEU over all languages doesn't correspond to training data sizes as Yoruba and Igbo; although one-fifth of the \texttt{Full} German data still achieves higher BLEU than German.

\section{Results for In-domain Validation and Test Sets}
\begin{figure*}
	\centering
	\vskip 0.15in
\begin{small}
\begin{sc}
	\begin{subfigure}{0.32\linewidth}
		\centering	\includegraphics[width=0.99\linewidth]{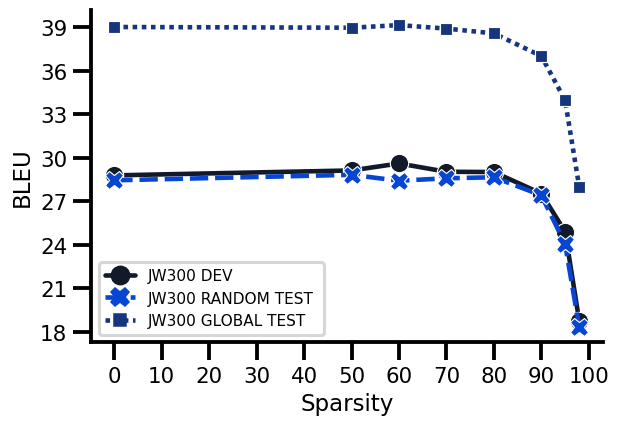}
    		\caption{BLEU (Yoruba)}
	\end{subfigure}
		\begin{subfigure}{0.32\linewidth}
		\centering
			
    	\includegraphics[width=0.99\linewidth]{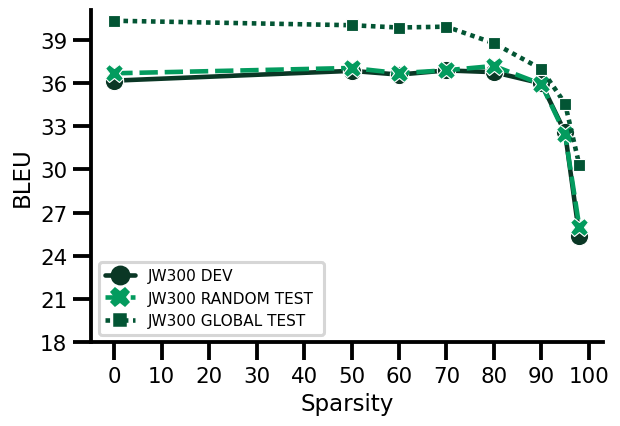}
    		\caption{ BLEU (Igbo)}
	\end{subfigure}
\\
	\begin{subfigure}{0.32\linewidth}
		\centering
    	\includegraphics[width=0.99\linewidth]{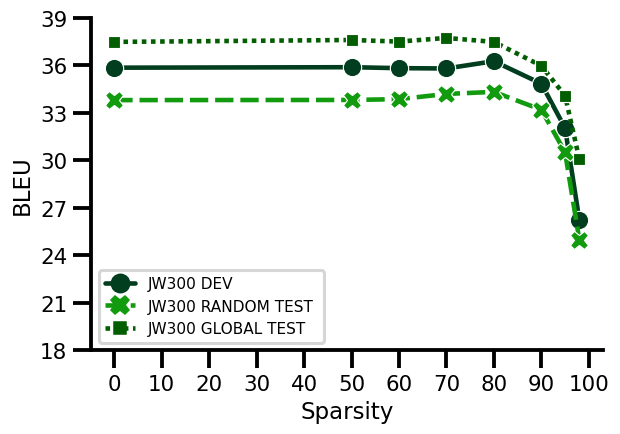}
    		\caption{BLEU (Hausa)}
	\end{subfigure}	
	\begin{subfigure}{0.32\linewidth}
		\centering	\includegraphics[width=0.99\linewidth]{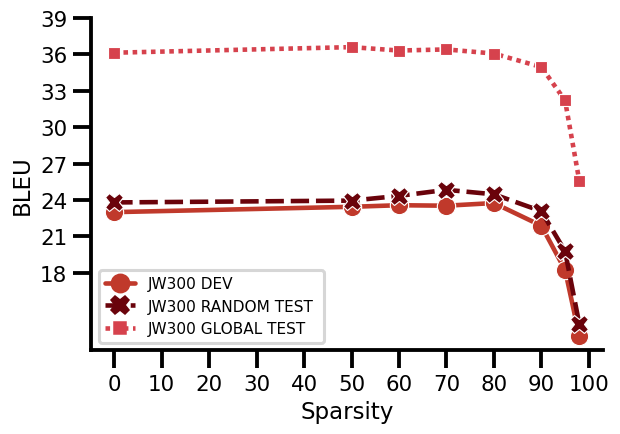}
\caption{BLEU (German)}
	\end{subfigure}	
	\end{sc}
	\end{small}
	\caption{
	   Comparison of Absolute BLEU performance of \texttt{Random Test}, \texttt{Global Test} and \texttt{Validation} sets
	}
	\label{fig:all_tests}
	\vspace{-0.9cm}
\end{figure*}

Figure~\ref{fig:all_tests} compares the performance across test and validation sets. We can see that the validation set is closer to the \texttt{Random} test set since it was sampled randomly as well. The \texttt{Global} test set, however, contains fewer long-tail examples and therefore receives higher BLEU.

\begin{table*}[th!]
\centering
\resizebox{2.0\columnwidth}{!}{
\begin{tabular}{ll|c|cc|cc|cc|cc} 
\toprule
Size & \%Sparse &\# Params & \multicolumn{2}{c|}{yo} & \multicolumn{2}{c|}{ig} & \multicolumn{2}{c|}{ha} &\multicolumn{2}{c}{de} \\
&  & & \texttt{Random} & \texttt{Global} & \texttt{Random} & \texttt{Global} & \texttt{Random} & \texttt{Global} & \texttt{Random} & \texttt{Global} \\
\midrule
Tiny & \texttt{0} & 14.0M & 20.39 & 30.11 & 28.79 & 31.78 & 27.74 & 31.84 & 15.40 & 28.05  \\
\midrule
\multirow{8}{*}{Base} &
\texttt{0} & 46.1M & 28.45 & 39.01 & 36.68 & \textbf{40.33} & 33.81 & 37.49 & 23.80 & 36.14 \\
& \texttt{50} & 23.1M &  \textbf{28.83} & 38.96 & 37.05 & 40.02 & 33.82 & 37.61 & 23.95 & \textbf{36.60} \\
& \texttt{60} & 18.6M &  28.42 & \textbf{39.14} & 36.69 & 39.86 & 33.87 & 37.51 & 24.34 & 36.32 \\
& \texttt{70} & 13.8M & 28.59 & 38.90 & 36.89 & 39.92 & 34.20 & \textbf{37.74} & \textbf{24.84} & 36.41 \\
& \texttt{80} & 9.4M &  28.66 & 38.57 & \textbf{37.19} & 38.77 & \textbf{34.33} & 37.49 & 24.47 & 36.07 \\
& \texttt{90} & 4.6M  & 27.42 & 37.01 & 35.93 & 36.93 & 33.20 & 32.05 & 23.08 &  34.96 \\
& \texttt{95} & 2.5M & 24.08 &  33.97 & 32.46 & 34.58 & 30.52 & 34.09 & 19.78 & 32.21\\
& \texttt{98} & 1.1M & 18.34 &  27.97 & 26.02 & 30.32 & 24.99 & 30.09 & 13.78 & 25.58\\
\bottomrule
\end{tabular}%
}
\caption{Number of non-zero parameters and test BLEU scores under the \texttt{Limited} training regime.}
\label{tab:parameter_count}
\vspace{-0.2cm}
\end{table*}

\vfill

\begin{table*}[h!]
\centering
\resizebox{1.6\columnwidth}{!}{
\begin{tabular}{ll|c|c|c|c|c} 
\toprule
Size & \%Sparse &\# Params & yo & ig & ha &de \\
&  & & \texttt{Global} & \texttt{Global} & \texttt{Global} & \texttt{Global} \\
\midrule
\multirow{8}{*}{Base} &
\texttt{0} &    46.1M & 39.01 & 40.33 & 37.49 & 36.14  \\
& \texttt{50} & 23.1M & 38.96 & 40.02 & 37.61 & 36.6    \\
& \texttt{60} & 18.6M & 39.14 & 39.86 & 37.51 & 36.32   \\
& \texttt{70} & 13.8M & 38.90 & 39.92 & 37.74 & 36.41   \\
& \texttt{80} & 9.4M &  38.57 & 38.77 & 37.49 & 36.07   \\
& \texttt{90} & 4.6M  & 37.01 & 36.98 & 35.98 & 34.96   \\
& \texttt{95} & 2.5M & 33.97 &  34.58 & 34.09 & 32.21   \\
& \texttt{98} & 1.1M & 27.97 &  30.52 & 30.09 & 25.58   \\
\bottomrule
\end{tabular}%
}
\caption{Number of non-zero parameters and test BLEU scores under the \texttt{Full} training regime.}
\label{tab:parameter_count_full}
\vspace{-0.2cm}
\end{table*}

\begin{table*}[th!]
\centering
\resizebox{2.0\columnwidth}{!}{
\begin{tabular}{l|c|c|c|cc|cc|cc|cc} 

\toprule

\%Sparse & BP & EP & PF & \multicolumn{2}{c|}{yo} & \multicolumn{2}{c|}{ig} & \multicolumn{2}{c|}{ha} &\multicolumn{2}{c}{de} \\

& & & & \texttt{Random} & \texttt{Global} & \texttt{Random} & \texttt{Global} & \texttt{Random} & \texttt{Global} & \texttt{Random} & \texttt{Global} \\
\midrule
\multirow{3}{*}{50} 
& 2000 & 60000 & 1000 & 28.45 & 38.82 & 36.93 & 39.247  & 33.81  & 38.14 & 24.26 & 35.77  \\
& 12000 & 40000 &  2000 & 28.83 & 39.07 & 37.55 & 39.91 & 33.95  & 37.9  & 24.52 & 36.6 \\
& 15000 & 40000 &  2000 & 28.64 & 38.77 & 37.05 & 40.02 & 34.02  & 37.99 &	23.95 & 36.25\\
\midrule
\multirow{3}{*}{60} 
& 2000  & 60000 & 1000  & 28.28 & 38.54 & 37.15 & 39.61 & 33.82 & 38.06 & 	24.16 & 36.09 \\
& 12000 & 40000 &  2000 & 28.42 & 39.14 & 36.69 & 39.86 & 33.68 & 37.9  &	24.3  & 36.32\\
& 15000 & 40000 &  2000 & 28.54 & 39.07 & 37.29 & 39.58 & 33.86 & 37.94 &   24.34 & 35.88\\
\midrule
\multirow{3}{*}{70} 
& 2000 & 60000 & 1000 & 28.12 & 38.46 & 37.55 & 38.88   & 34.10 & 37.84 & 24.54 & 36.2	\\
& 12000 & 40000 &  2000 & 28.46 & 38.6 & 36.89 & 39.77  & 33.87 & 38.29 & 24.84 & 36.41	\\
& 15000 & 40000 &  2000 & 28.59 & 38.9 & 37.32 & 39.92  & 33.80 &  38.15 & 24.41 & 36.25	      \\
\midrule
\multirow{3}{*}{80} 
& 2000 & 60000  & 1000  & 28.36 & 38.07 & 36.73 & 38.36 & 33.72 & 37.52 & 23.83 & 36.02	 \\
& 12000 & 40000 & 2000  & 28.66 & 38.57 & 37.1 & 38.54  & 34.41 &  37.95 & 24.47 & 36.07 \\
& 15000 & 40000 &  2000 & 28.53 & 38.19 & 37.19 & 38.77 & 34.33 & 38.03  & 24.45 & 35.82 \\
\midrule
\multirow{3}{*}{90} 
& 2000 & 60000 & 1000 & 26.50 & 36.09 & 35.42 & 36.73 & 32.35 & 36.38 & 22.68 & 34.29 \\
& 12000 & 40000 &  2000 & 27.42 & 37.01 & 35.93 & 36.98 & 33.20 & 36.51 & 23.08 &  34.96 \\
& 15000 & 40000 &  2000 & 26.96 & 36.72 & 36.08 & 36.91 & 33.15 & 36.41 & 23.23 &  34.36 \\
\midrule
\multirow{3}{*}{95} 
& 2000  & 60000 &   1000 & 24.08 & 32.82 & 31.49 & 34.06 & 29.16 & 33.32 &	18.97 & 31.5\\
& 12000 & 40000 &   2000 & 23.93 & 33.71 & 32.46 & 34.58 & 30.69 & 34.39 & 	20.01 & 32.14 \\
& 15000 & 40000 &   2000 & 24.12 & 33.06 & 32.44 & 34.29 & 30.52 & 34.61 & 19.78 & 32.21\\
\midrule
\multirow{3}{*}{98} 
& 2000  & 60000 &   1000 & 18.34 & 25.95 & 25.05 & 28.87 & 24.22 & 28.63 & 12.9 & 24.35	 \\
& 12000 & 40000 &   2000 & 17.87 & 27.47 & 26.02 & 30.32 & 24.99 & 30.09 & 14.25 & 25.58	\\
& 15000 & 40000 &   2000 & 17.54 & 27.28 & 26.00 & 29.39 & 24.99 & 29.5 & 13.78 & 25.13	 \\
\bottomrule
\end{tabular}%
}
\caption{Absolute BLEU performance from tuning pruning hyperparameters across all languages under the \texttt{Limited} training regime. \textbf{BP}= Begin Pruning Step, \textbf{EP}= End Pruning Step and \textbf{Pruning Frequency}= Pruning Frequency}
\label{tab:tuning_hparams_samp}
\end{table*}

\end{document}